\documentclass[runningheads]{llncs}
\usepackage{graphicx}
\usepackage{amsmath,amssymb,amsfonts,bm}
\usepackage{algorithmic}
\usepackage{subcaption}
\usepackage{booktabs}
\usepackage{multirow}
\usepackage[ruled,norelsize,linesnumbered]{algorithm2e}

\usepackage[colorlinks=true,urlcolor=black]{hyperref}
\begin{document}
%
\title{Gradient-based Data Subversion Attack Against Binary Classifiers}
%
%
\author{Rosni K Vasu \and Sanjay Seetharaman \and Shubham Malaviya \and Manish Shukla \and Sachin Lodha}

\authorrunning{Vasu et al.}
%
\institute{TCS Research, India\\
\email{\{rosni.kv,s.seetharaman1,shubham.malaviya,mani.shukla,sachin.lodha\}@tcs.com}}
\maketitle              
\begin{abstract}
	Machine learning based data-driven technologies have shown impressive performances in a variety of application domains. Most enterprises use data from multiple sources to provide quality applications. The reliability of the external data sources raises concerns for the security of the machine learning techniques adopted. An attacker can tamper the training or test datasets to subvert the predictions of models generated by these techniques. Data poisoning is one such attack wherein the attacker tries to degrade the performance of a classifier by manipulating the training data.
	
	In this work, we focus on label contamination attack in which an attacker poisons the labels of data to compromise the functionality of the system. We develop Gradient-based Data Subversion strategies to achieve model degradation under the assumption that the attacker has limited-knowledge of the victim model. We exploit the gradients of a differentiable convex loss function (residual errors) with respect to the predicted label as a warm-start and formulate different strategies to find a set of data instances to contaminate. Further, we analyze the transferability of attacks and the susceptibility of binary classifiers. Our experiments show that the proposed approach outperforms the baselines and is computationally efficient.

\keywords{adversarial attacks \and data poisoning attacks \and label contamination, transferability of attack \and susceptibility.}
\end{abstract}

\section{Introduction} \label{sec:introduction}
In last decade machine learning techniques have been extensively used in variety of domains, for example, malware detection, object detection, natural language processing and recommender system. However, in-spite of its success, it is still vulnerable to different kind of adversarial attacks. Input data poisoning is one such adversarial attack, wherein an adversary who is capable of accessing the training data could easily contaminate it. A model which is trained using this poisoned data is corrupted and, thus produces unexpected results after deployment. Data poisoning attack is a serious threat as it can result in security and privacy issues, economic losses, and has ethical and legal implications.

Biggio et al \cite{biggio2012poisoning} defined data poisoning attack as a causative attack in which training data is injected with specially crafted attack points. The attacker can have direct access to the training database (insider threat) \cite{homoliak2019insight} or no access but may provide new training data \cite{biggio2012poisoning}. Anti-virus vendors have been previously blamed for injecting poisoned samples into VirusTotal\footnote{ https://www.virustotal.com/} for degrading the performance of the competing products \cite{biggio2018wild}. The machine learning models are more susceptible to data poisoning attacks when the training data is coming from an unconstrained and unmonitored source, for example crowd. For example, the twitter bot, \emph{Microsoft Tay}, was designed to engage with users of age 18 to 24, however, within 16 hours the bot started posting inflammatory and offensive tweets by analyzing and updating its model based on the interactions with people on Twitter \cite{price2016microsoft}. Similarly, Lam et al \cite{lam2004shilling} studied the shilling attack in recommender systems wherein the untrustworthy participants could influence the system in recommending low-quality items to an unsuspecting user. 

Previous work in adversarial machine learning has shown that a malicious adversary could stealthily manipulate the input data to compromise the security of the machine learning system \cite{goodfellow2018making}. Some well-studied attack strategies in the existing literature includes minimum-distance evasion of linear classifiers, gradient-based adversarial perturbations \cite{biggio2018wild}, projected gradient descent attack \cite{zhao2017efficient} and maximum-confidence evasion \cite{demontis2018securing}. Further, Lowd and Meek \cite{lowd2005good} have shown that the spam classifiers can be easily fooled by carefully crafting the content of the spam emails without affecting their readability. Similarly, researchers have shown that it is easy to bypass anti-malware solutions by either modifying the API call sequence of the malware \cite{rosenberg2018generic} or the structure \cite{maiorca2013looking}.

Two of the major attacking strategies are \emph{evasion} and \emph{poisoning} attacks. In an evasion attack, the test data is manipulated to evade the classifier's decision boundary. In a recent work, Labaca et al \cite{labaca2019poster} demonstrated the evasion attack on malware classifiers, wherein they generated valid adversarial malware samples against convolutional neural networks using gradient descent technique. In contrast to this, in a data poisoning attack, the training data is manipulated to achieve the attacker's objective. For example, Nelson et al \cite{nelson2008exploiting} have shown a targeted attack on a spam filter, which makes the system useless even when the adversary modifies only 1\% of training data. In this work we primarily focus on the data poisoning attack.

\emph{Label contamination attack} is an example of input data poisoning attack, in which the adversary introduces label noise in the training data. Previous work on label contamination have demonstrated attack on the Support Vector Machines (SVM) with an assumption that the attacker has full knowledge of the target model \cite{biggio2011support,xiao2012adversarial,biggio2012poisoning}. Zhao et al \cite{zhao2017efficient} extended the attack strategies to other linear classifiers (Logistic Regression and Least-Square SVM). Moreover, they have demonstrated the transferability of attacks in a black-box setting where the victim's learning model is unknown to the attacker.

In this work, a gradient-based data subversion attack is presented with primary focus on label contamination against binary classifiers. Also, we propose an algorithm to find an optimal set of data samples from the training data to poison. Additionally, our proposed method assumes that the attacker only knows about the input feature representation (black-box setup). Our proposed attack strategy is generic and is applicable to multiple application types, for example, spam filters, intrusion detection, object classification, and stock prediction. Following are the major contributions of our paper:

\textbf{\emph{Search Space Reduction.}} We propose an efficient label contamination strategy which is warm-started with the gradients of a differentiable convex loss function (residual error) which helps in search space reduction for finding optimal data poisoning instances.

\textbf{\emph{Efficient Attack Strategy.}} he cost of finding the set of near-optimal label flips is directly proportional to the number of training instances. Hence, the attacker has to perform expensive operations with enormous training data. OGDS solves one linear-programming problem and retrains the classifier in each iteration. This reduces the search space of the data instances to poison, which makes it efficient compared to the existing label contamination approaches. Please refer to Section \ref{sec:ComplexityAnalysis} (Complexity Analysis) for more details.

\textbf{\emph{Varied Cost Analysis.}} The motivation behind considering varied cost in data input poisoning is that not all data-points affect the model equally. Some data-instances, if poisoned, have more detrimental effect on the performance of the model. Refer to Section \ref{sec:cost} (Analysis of Cost Function) for more details.

\textbf{\emph{Relaxed Assumptions About the Attacker.}} We have assumed a black-box setting; hence the attacker has limited information. Further, we have demonstrated the effectiveness of our attack on multiple datasets from different domains.

The rest of the paper is structured as follows. We discuss our threat model in Section \ref{sec:threat} and related work in Section \ref{sec:related works}. We define the data poisoning framework in Section \ref{sec:DP} and describe the background information on the gradient-based decision tree in Section \ref{sec:prelim}. We introduce the gradient-based data subversion framework in Section \ref{sec:GDS}. The experimental setup and further analysis are reported in Section \ref{sec:experiments}. The susceptibility of models is defined and discussed in Section \ref{sec:susceptibility}. We discuss insights on possible defense mechanisms and attack scenarios in Section \ref{sec:defense} and conclude with discussion in Sections \ref{sec:discussion} and \ref{sec:conclusion}.

\section{Threat Model and Assumptions}\label{sec:threat} In general, the attacker can either cause \emph{integrity} violation where the normal system functionality is maintained while evading the classifier or \emph{availability} violation where the normal system functionality is compromised thereby making it unavailable for legitimate users \cite{xiao2015feature,zhao2017efficient}. We focus on the \emph{availability attack} where the attacker's goal is to degrade the performance of the system.

The attacker's knowledge of the victim system can include the training data, the feature set, the learning algorithm along with the objective function minimized during training; and, possibly, it's (trained) parameters/hyper-parameters $\mathbf{w}$ \cite{biggio2018wild}. The variation in the attacker’s knowledge leads to attacks ranging from white-box to black-box. The attack can be categorized as a \emph{white-box} when the attacker has full knowledge of the target classifier. In contrast, the \emph{black-box} attack is when the attacker has limited knowledge of the components. 

We will focus on the context where the adversary, a malicious data provider, has access to the training data. This threat is important and well known for the security of machine learning classifiers as many data-driven solutions rely on the external data sources.

The attacker's ability defines how the attacker can influence the system. Based on the ability of the attacker to manipulate both the training and test or only the test data, the attack is said to be causative (poisoning) and exploratory (evasion) respectively \cite{demontis2018securing,biggio2012poisoning,xiao2012adversarial}. In our work, we focus on the \emph{causative availability attack}. 

The existing data poisoning frameworks either perturb the input instances or contaminate the labels. In some application domains such as malware detection, maintaining the integrity of the input instances after adversarial perturbation requires domain knowledge and effort. We restraint the capability of the attacker such that the attacker can only flip the labels of the data. Moreover, we assume that the attacker has limited knowledge of the victim model such as only the training data, input feature representation.

In our work, the adversarial budget $\mathcal{B}$ is defined as the number of flips. The budget can also be: 1. sum of monetary value associated with each flip 2. system applied constraints like the number of queries that attacker can make 3. cost associated with the attacker's time limit and the sequential nature of data. In this work, we devise an algorithm to compute the label contamination attack under a given budget. 

\section{Related Work} \label{sec:related works}
Studies on the label contamination attacks have been more focused on Support Vector Machines (SVM) and are not generalized to other classifiers \cite{biggio2011support,biggio2012poisoning,xiao2012adversarial}. The attack strategy by Biggio et al in \cite{biggio2011support} flips the labels of the data points that are non-support vectors and decreases the probability of contaminating the labels of support vectors. Also, they demonstrated that a random label flip of 40\% of data samples in a synthetic dataset can deteriorate the performance of SVMs by reducing their accuracy ($\approx$25\%). They further crafted inputs using a gradient ascent method and inserted them into the training sets and degraded the performance of the classifier \cite{biggio2012poisoning}. Xiao et al \cite{xiao2015feature} proposed the attacks on feature selection algorithms and investigated their robustness. 

Xiao et al \cite{xiao2012adversarial} proposed the approach ALFA for finding the near-optimal label flips and compared it with uniform random flip, nearest-first flip (minimum distance to the decision boundary) and furthest-first flip (maximum distance to the decision boundary) strategies. Moreover, they assumed that the attacker has full knowledge of the victim's learning model. This assumption considered as the starting point for understanding the attack strategies \cite{mei2015using}. Paudice et al \cite{paudice2018label} used a heuristic approach which aims to find the required number of instances that maximizes the error function of the model on a validation set. Further, the authors proposed a defense strategy that re-labels the points that are suspicious to be malicious based on the nearest neighbor approach. 

Mei et al \cite{mei2015using} formalized the training set attack as a bilevel optimization framework. Their objective was to modify the training data such that the distance between the learner's learned model and the original model is minimum, in addition to satisfying the attacker's objectives using the least number of manipulations. They demonstrated their strategy on the SVM, Logistic and Linear Regressions. Zhao et al \cite{zhao2017efficient} studied the label contamination attack as a mixed integer bilevel optimal problem and proposed \emph{Projected Gradient Ascent} (PGA) algorithm to solve it. Their approach was more realistic compared to prior-arts as they considered attackers with arbitrary objectives in a black-box setting. Also, they studied the transferability of poisoning attacks.

Transferability of data poisoning attacks has been studied in prior works \cite{munoz2017towards,suciu2018does,transferUsenix}. Mu\~{n}oz-Gonz\'{a}lez et al developed a data poisoning attack for deep networks \cite{munoz2017towards} and demonstrated the transferability of their attack across different learning algorithms. Demontis et al \cite{demontis2018securing,transferUsenix} studied the transferability of evasion and poisoning attacks in their recent work. These studies shed some insights on the factors that can influence the attack transfer. In our work, we show the transferability of our attack in various learning algorithms.

The existing label contamination attack strategies analyze all the training instances for finding a set of near-optimal data instances to flip. They solve a bilevel optimization problem \cite{xiao2012adversarial,biggio2011support,mei2015using,zhao2017efficient} which is complex in terms of solving two optimization problems within each iteration. The attack strategy PGA \cite{zhao2017efficient} trains a linear classifier in each iteration which is more efficient than the former. However, they consider the dual form of the classifiers that increases the number of computations when the training set is large. 

Koh et al \cite{koh2017understanding} efficiently approximated influence functions which formalize the impact of a training point on a machine learning model. They computed the influence by forming a quadratic approximation to the empirical risk around model parameters and taking a single newton step. Since we compute the label flip attack on a given dataset, we filter the points based on the gradients of the loss function with respect to the predicted label. 

Recent studies have developed data poisoning attacks and defenses for online learning \cite{wang2018data,zhang2019online,wang2019investigation}. In this setting, attacker has to consider the online nature of the problem where data is sequential in nature and the classifier is a more complex function of the data stream. Moreover, the order of the data influences the attack. Further, the knowledge about the future data point is limited for the attacker. All of these together makes the online learning attack more challenging. We consider the offline setting in this work and propose an algorithm to find the candidate set by filtering the points from the training instances. We evaluate our approach in multiple datasets. Further, the transferability of the proposed algorithm on various machine learning models is studied.
\begin{figure}[ht]
	\centering
	\includegraphics[width=\textwidth]{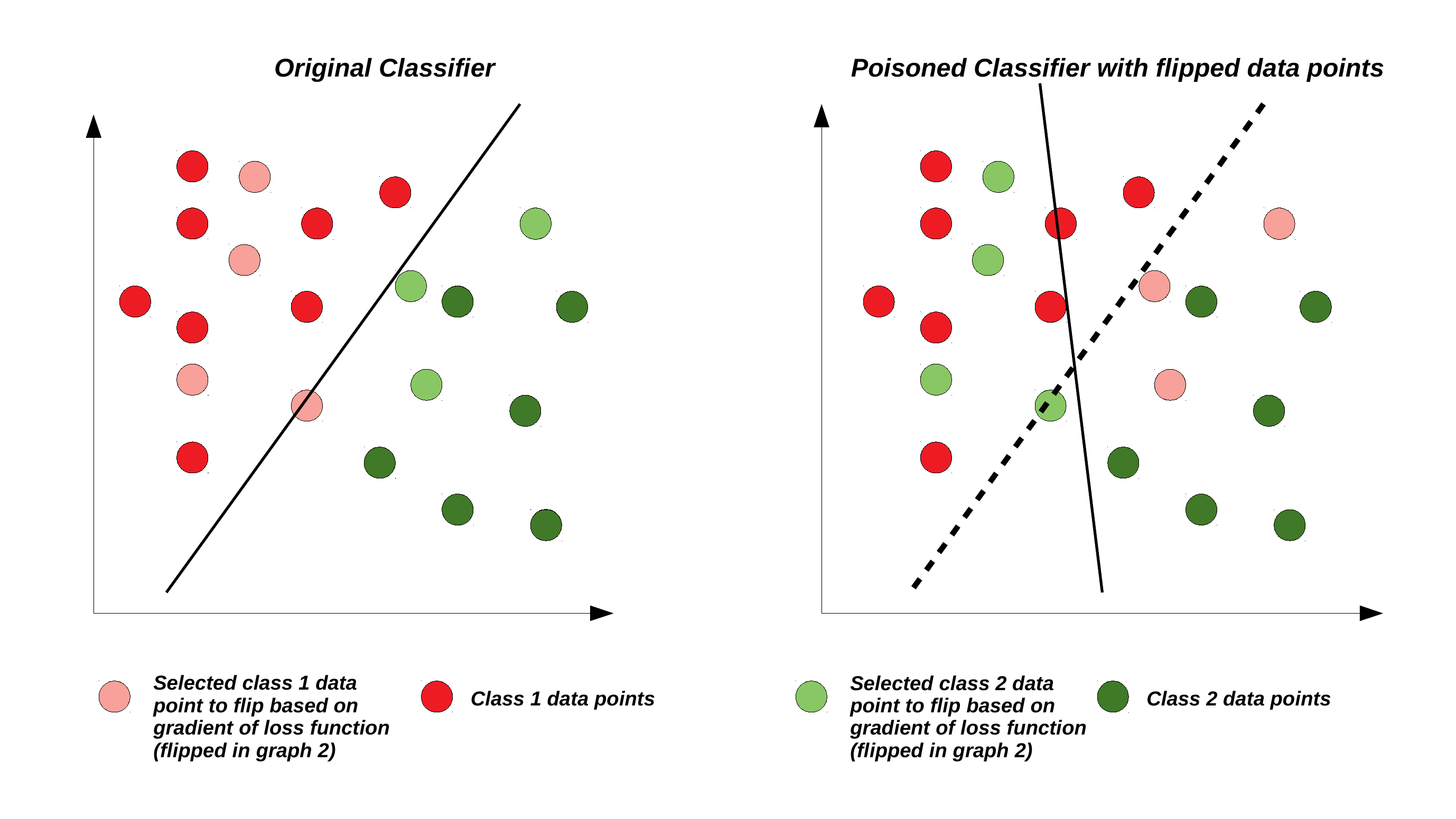}
	\caption{Visualization of the Data Poisoning Attack. \emph{Light green} and \emph{light red} data samples in the first graph are selected based on the gradients. In the second graph, we can see that an adversary who is flipping the labels of those data points can change the decision boundary of the classifier. The \emph{dotted line} is the original decision boundary and the \emph{solid line} indicates the poisoned decision boundary.}
	\label{fig:vizConcept}
\end{figure}

\section{Data Poisoning Attack Framework} \label{sec:DP}
Consider a supervised learning problem with training data $\mathcal{D}$ = $\{(\mathbf{x}_{i},y_{i})\}_{i=1}^{n}$, where $\mathbf{x}_{i} \in \mathbb{R}^{d}$ is the $i^{th}$ input instance and $y_{i} \in \{-1,1\}$ is the corresponding classification label. For a linear classifier, the objective function is 
\begin{equation}
	\begin{aligned}
		{\underset{f \in \mathcal{H}}{min}} \; \gamma \sum_{i=1}^{n} \mathcal{V}(y_{i},f(\mathbf{x}_{i})) + \frac{1}{2} {\parallel \mathbf{w} \parallel}^{2}
		\label{eq:ObjFun}
	\end{aligned}
\end{equation}

where $f(\mathbf{x}_{i}) = \mathbf{w}^{T}\mathbf{x}_{i} + b$ is the decision function learned by minimizing the objective function (Equation \ref{eq:ObjFun}) on the training data $\mathcal{D}$, ${\parallel \mathbf{w} \parallel}^{2}$ is the square of the $\ell_{2}$ norm of the weight vector $\mathbf{w}$, $\mathcal{H}$ is the hypothesis space, $\mathcal{V}$ is the loss function and $\gamma$ is the regularisation parameter. For a given test instance $\mathbf{x}_{i}$, $\hat{y}_{i}= sgn(f(\mathbf{x}_{i}))$ is the predicted label.

We perform an \emph{error-generic} poisoning attack \cite{munoz2017towards} so that the victim classifier will have maximal classification error on some test data. We select a subset of training instances to flip, under the given budget. That is, the set of poisoned instances $\mathcal{D}_{p}$ is chosen such that it has maximal loss under the original classifier $f_{\mathcal{D}}$ and minimal loss under the poisoned classifier $f_{\mathcal{D}_{p}}$ \cite{xiao2012adversarial}.

The poisoning attack requires solving a bilevel optimization problem in order to find the near-optimal label flips. It can be formulated as a loss minimization framework \cite{munoz2017towards,xiao2012adversarial,biggio2012poisoning}. For $i=1,\dots,n$ and $j=n+i$, we construct $\mathbf{x}_{j}=\mathbf{x}_{i},y_{j}=-y_{i}$ such that $ S = \{(\mathbf{x}_{i},y_{i})\}_{i=1}^{n} \cup \: \{(\mathbf{x}_{j},y_{j})\}_{j=n+1}^{2n}$. 

\begin{equation}
	\tag{IP-I}
	\begin{aligned}
		{\underset{q,f}{min}} \; \gamma \sum_{i=1}^{2n} \; & q_{i} [\mathcal{V}(y_{i},f_{\mathcal{D}_{p}}(\mathbf{x}_{i})) - \mathcal{V}(y_{i},f_{\mathcal{D}}(\mathbf{x}_{i})) ]  + {\parallel \mathbf{w} \parallel}_{\mathcal{H}}^{2} \\
		& \textbf{s.t. } \sum_{i=n+1}^{2n} c_{i}q_{i} \le \mathcal{B} \\ 
		& q_{i} + q_{i+n} = 1, \; i=1,\dots,n \\
		& q_{i} \in \{0,1\}, \; i = 1,\dots,2n
		\label{eq:lossMin}
	\end{aligned}
\end{equation} 

where, ${\parallel \mathbf{w} \parallel}_{\mathcal{H}}^{2}$ is constant with respect to the optimization variables, $q$ is the indicator variable for each instance in $S$. Each indicator variable $q_{i} = 1$ implies $(\mathbf{x}_{i},y_{i}) \in \mathcal{D}_{p}$ and $q_{i}=0$ indicates that it is not in the poisoned dataset. Also, $q_{i} + q_{i+n} = 1$ constrain the poisoned training set such that instance $\mathbf{x}_{i}$ can be selected with only one label. The total adversarial budget is given by $\mathcal{B}$ and $c_{i} \in \mathbb{R}_{+}$ is the cost of contaminating a single label $y_{i}$ \cite{xiao2012adversarial}. The optimization problem of the decision function $f(.)$ within \ref{eq:lossMin} makes it a bilevel formulation. But this procedure is time consuming, especially when working with large datasets.

\section{Search Space Reduction} \label{sec:prelim}
Each data instance in the training data influences the training process differently. We leverage this fact to speed up the process of solving the near-optimal label flip problem. The original dataset is reduced to a smaller dataset that contains the most influential points. This helps us to reduce the search space and solve the problem much faster for large datasets. The solution obtained is almost as good as the original solution (considering the complete dataset) because the filtering procedure may potentially exclude points that were part of the original solution. Therefore, this reduction quickens the process of solving \ref{eq:lossMin} while achieving almost the same degradation on the victim model.

We develop a procedure to sample the most influential data instances based on the work by Ke et al \cite{ke2017lightgbm}. For their sampling technique, they computed the gradients of the loss function with respect to the output of the model and noticed that data instances with different gradients played different roles in the computation of information gain. In particular, the data instances with small gradient value have small training error and are well-trained. They proposed a sampling technique based on the absolute values of the gradients of the data instances. This technique sorts the instances based on the gradients and selects a fraction of instances with the largest gradients. Then it randomly samples another fraction of instances from the rest of the data so that the original data distribution is maintained.

We first compute the gradients of all instances in the training set and pick the ones with small gradients. In order to maintain the data distribution, we randomly sample a small fraction of instances with large gradients. We find that the set of top \emph{b} $\times$ 100\% instances with small gradients and \emph{a} $\times$ 100\% instances with large gradients is a good candidate for label contamination, with $0 \le a,b,a+b \le 1$. We thus exclude a significant proportion of data instances to mount the attack. This help us to reduce the search space of data instances without impeding the attack performance.

\subsection{Gradient Computation} 

We compute the gradients based on Gradient Boosted Decision Tree (GBDT), a popular and practical machine learning algorithm because of its efficiency, accuracy and interpretability \cite{friedman2001greedy,chen2016xgboost}. For a given data instance $\mathbf{x}_{i}$, the prediction value ($\hat{y_{i}}$) for a tree ensemble model uses \emph{K} additive functions
\begin{equation}
	\hat{y}_{i} = \sum_{m=1}^{M} f_{m}(\mathbf{x}_{i})
	\label{eq:EnsembleModel}
\end{equation}
where, each $f_{m}$ is a decision tree. Chen et al \cite{chen2016xgboost} defined the tree ensemble by a vector of scores in leafs and, a leaf index mapping function maps an instance to a leaf.
\begin{equation}
	f_{m}(\mathbf{x}) = w_{s(\mathbf{x})}, \; w \in \mathbb{R}^{T},\: s: \mathbb{R}^{d}
	\rightarrow \{1,2, \dots, T\}
	\label{eq:tree}
\end{equation}
where, $w$ is the leaf weight of the tree, $s$ is structure of the tree which maps an instance to the leaf index and T indicates the number of leaves in the tree.

The simplified objective function ($\mathcal{L}$) at $t^{th}$ iteration given in Section 2.2 of \cite{chen2016xgboost} is,
\begin{equation}
	\mathcal{L}^{(t)} = \sum_{i=1}^{n} [g_{i}f_{t}(\mathbf{x}_{i}) + \frac{1}{2} h_{i}f_{t}^{2}(\mathbf{x}_{i})] + \Omega(f_{t})
	\label{eq:xgboost}
\end{equation}
\begin{equation}
	g_{i} = \partial_{\hat{y}^{(t-1)}} l(y_{i},\hat{y}^{(t-1)})
	\label{eq:g}
\end{equation}
\begin{equation}
	h_{i} = \partial^{2}_{\hat{y}^{(t-1)}} l(y_{i},\hat{y}^{(t-1)})
\end{equation}
where, $g_{i}$ is the gradient value and $h_{i}$ is the hessian value from the differentiable convex loss function ($l$) wherein $l$ computes the difference of predicted ($\hat{y}_{i}$) and actual value ($y_{i}$), $\Omega$ is the regularisation term to avoid overfitting.

\section{Gradient-based Data Subversion Attack} \label{sec:GDS}

In this work, we consider causative attacks which degrade the classifier performance by poisoning the training data. Similar to previous studies \cite{biggio2012poisoning,paudice2018label,zhao2017efficient}, we consider a binary classification problem with the objective of maximizing the error of victim classifier (Refer \ref{eq:lossMin}). Moreover, we assume that the attacker has limited-knowledge of the victim system.

\subsection{Our Attack Strategy} We leverage the absolute value of gradients to reduce the search space of data instances to poison. We have the training data $\mathcal{D}$ and a set of validation data samples with \emph{t} instances, $\mathcal{D}_{test} = \{(\mathbf{x}_{i},y_{i})\}_{i=1}^{t}$ with an assumption that the data in both the sets are genuine. The large gradient values contribute to the computation of \emph{information gain} in the decision tree, which is then used to find the split point of the tree. Our approach considers different proportions of large and small gradient data instances for label contamination. An illustration of the concept is shown in the Figure \ref{fig:vizConcept}. The complete procedure is given in Algorithm~\ref{algo:GDS}.

    \begin{algorithm}[ht]
        \KwIn{Original training data D = $\{(\mathbf{x}_{i},y_{i})\}_{i=1}^{n}$, budget B, gradient values $\nabla = \{g_{1},\dots,g_{n}\} $, sampling ratio a and b, iteration limit $t_{max} $ }
        
        $\Gamma \leftarrow$ Indices of Sort($\nabla, \textit{`desc'} )$ \\
        
        $ \nabla_{\Gamma(b)} = \nabla_{\Gamma[- b \ast \lvert D \rvert:]}$ \tcp*{Indices of top b \%}
        
        $ \nabla_{\Gamma(a)} = \nabla_{\Gamma[randA]}$ \tcp*{Indices of a \%, randA = randPick$(\{\Gamma - \Gamma(b)\},a \ast \lvert D \rvert)$}
        
        $ \nabla_{\Gamma(data)} = \nabla_{\Gamma(a)} + \nabla_{\Gamma(b)} $ \tcp*{Indices of data to be flipped, $ \lvert \nabla_{data} \rvert = $ k instances}
        
        t $ \leftarrow $ 1 \\
        
        \While{t $\le t_{max} $}
        {
            $ \nabla_{ind}^{(t)} \leftarrow $ Choose a random $ \nabla_{ind} \in [0,1]^{k} $ 
            \tcp*{ $ \nabla_{ind} $ maps to indices of $\nabla_{data}$ } 
            
            $y^{'(t)} \leftarrow $ Flip ($\nabla_{ind}^{t}$)
            
            $ f^{(t)}(\mathbf{x}) \leftarrow $ Retrain the classifier using data $\{(\mathbf{x}_{i}, y^{'}(t))\}$ 
            
            e (t) $\leftarrow \mathcal{V}(y, f^{(t)}(\mathbf{x}_{test}))$
            \tcp*{ $ \mathcal{V} $ is the loss function of the classifier}
        }
        $ t_{f} = \underset{t}{\mathbf{argmax}}$ e \\
        
        $\mathcal{D}_{p} =  \{ (\mathbf{x}, y^{'}(t_{f}))\}_{1}^{n} $ \\
        
        \KwOut{Poisoned training data $\mathcal{D}_{p}$ }
        
        \caption{Gradient-based Data Subversion (GDS)}
        \label{algo:GDS}
    \end{algorithm}

The data is sorted based on their gradients ($\nabla$) which are computed beforehand (refer to Equation \ref{eq:g}). Based on the sampling ratios we have \emph{k} instances, where \emph{k}$<$\emph{n}. Now, we have to find the data samples to poison from this relatively smaller dataset. To the best of our knowledge, the strategy of finding relevant data instances to poison using gradients from boosting models is not considered in any of the previous works.

We choose random indicators uniformly for this \emph{k} instances, such that $\nabla_{ind}(i)$ = 1 indicates that label should be flipped, otherwise it is not. \emph{Budget} is defined as the adversarial cost for contaminating data instances. Here, we consider the budget $\mathcal{B}$ as the number of instances an adversary can flip in order to poison the model \cite{xiao2012adversarial}. Based on the budget, we flip at most $\mathcal{B}$ instances randomly (Algorithm \ref{algo:Flip}).

The flipped data is then used to retrain the classifier. The error value is computed and saved in $\mathbf{e}$ using the loss function $\mathcal{V}$. We continue this for $t_{max}$ iterations. We choose the poisoned data $\mathcal{D}_{p}$ which is giving maximum error on the validation data (Algorithm \ref{algo:GDS}).

\begin{algorithm}[ht]
    \KwIn{ $\nabla_{ind}$, Original labels y, budget $\mathcal{B}$, $\Gamma$ }
    $ \mathbf{y}^{'} \leftarrow \mathbf{y}$ \\
    $ j \leftarrow 1 $ \\
    $sum \leftarrow \nabla_{ind}[j] $ \\
    \While{$ sum \leq \mathcal{B} $}
    {
        $y^{'}_{\Gamma(j)} \leftarrow - y_{\Gamma(j)} $ \\ 
        $ j \leftarrow j + 1 $	\\	
        $ sum \leftarrow sum + \nabla_{ind}[j]$
    }
    \KwOut{ Flipped labels $y^{'}$}
    \caption{Flip Function, Flip(.)}
    \label{algo:Flip}
\end{algorithm}

\begin{algorithm}[ht]
	\KwIn{Original training data D = $\{(\mathbf{x}_{i},y_{i})\}_{i=1}^{n}$, budget B,  gradient values $\nabla = \{g_{1},\dots,g_{n}\} $, sampling ratio a and b, iteration limit $t_{max} $, adversarial cost $c_{1},\dots,c_{n}$ }
	
	$\Gamma \leftarrow$ Indices of Sort($\nabla, \textit{`desc'} )$ \\
	$ \nabla_{\Gamma(b)} = \nabla_{\Gamma[- b \ast \lvert D \rvert:]}$ \tcp*{Indices of top b \%}
	$ \nabla_{\Gamma(a)} = \nabla_{\Gamma[randA]}$ \tcp*{Indices of a \%, randA = randPick$(\{\Gamma - \Gamma(b)\},a \ast \lvert D \rvert)$}
	$ \nabla_{\Gamma(data)} = \nabla_{\Gamma(a)} + \nabla_{\Gamma(b)} $ \tcp*{Indices of data to be flipped, $ \lvert \nabla_{data} \rvert = $ k instances}
	$ f^{(0)}(\mathbf{x}) \leftarrow $ Train original classifier using data $\{(\mathbf{x}_{i}, y_{i})\}$ \\
	\bm{$\epsilon$} $ \leftarrow $ $[0]_{0}^{2k}$
	Compute \textbf{e} using $ f^{(0)}(\mathbf{x})$ \tcp*{classifier error, $\lvert$\textbf{e}$\rvert$ = 2k}
	t $ \leftarrow $ 1 \\
	
	\While{ not converge and t $\le t_{max} $}
	{
		Find \textbf{q} = [$q_{1},\dots,q_{2k}$] by solving \ref{eq:minq2} \\
		Obtain $y^{'}(t)$ using \textbf{q} \\
		$f^{t}(\mathbf{x}) \leftarrow $ Retrain the classifier using the training data $ \{ (\mathbf{x}_{i},y^{'}(t))\} $ \\
		Compute \bm{$\epsilon$} using $f^{t}(\mathbf{x})$ \\
		Error (t) $\leftarrow \mathcal{V}(y_{test}, f^{(t)}(\mathbf{x}_{test}))$ \\
		$t \leftarrow t + 1$ \\		 
	}	
	$ t_{f} = \underset{t}{\mathbf{argmax}}$ Error \\
	$\mathcal{D}_{p} =  \{ (\mathbf{x}, y^{'}(t_{f}))\}_{1}^{n} $ \\				
	\KwOut{Poisoned training data $\mathcal{D}_{p}$ }
	\caption{Optimized Gradient-based Data Subversion (OGDS)}
	\label{algo:OGDS}
\end{algorithm}

\subsection{Optimized Gradient-based Data Subversion Attack (OGDS)} \label{sec:ogds}
Inspired by the attack on SVM \cite{xiao2012adversarial}, we formulate OGDS using an integer linear programming problem based on Equation \ref{eq:lossMin}. The complete approach is described in Algorithm \ref{algo:OGDS}.

In this algorithm, we start with an approach similar to GDS by sorting the gradient values and choosing \emph{k} instances. We reduce the search space from  the input of $\mathcal{D}^{n}$ to $\mathcal{D}^{k}$, where \emph{k}$<$\emph{n}. Consider an indicator variable $\mathbf{q}$ for the k data instances (refer the Equation \ref{eq:lossMin}). By solving the integer linear programming problem (ILP), Equation \ref{eq:minq}, we can have the $\mathbf{q}$ that minimizes the objective function and satisfies the budget constraint.
\begin{equation}
	\tag{IP-II}
	\begin{aligned}
		\underset{\mathbf{q}}{min} \; & \sum\limits_{i=1}^{2k} \; q_{i} (\epsilon_{i} - e_{i})  \\
		& \textbf{s.t.} \sum\limits_{i=k+1}^{2k} c_{i} q_{i} \le \mathcal{B} \\
		& q_{i} + q_{i+k} = 1,\; i = {1,\dots,k} \\
		& q_{i} \in \{0,1\}, \; i = {1,\dots,2k}
	\end{aligned}
	\label{eq:minq}
\end{equation}
where, error value $\epsilon_{i}$ computed from previous iteration's model $f^{(t-1)}(\mathbf{x})$ and error value $e_{i}$ computed from original classifier f($\mathbf{x}$). In this work, we set the cost for each label flip as one. We conjecture that the attacker's choice of the cost function is insignificant and analyze it in section \ref{sec:cost}. 

The integer linear programming problem (Refer \ref{eq:minq}) is NP-Hard. Hence, we perform LP relaxation on it and remove the integrality constraints. This makes $q_{i}$ a continuous variable which can take any real value over $[0,1]$.

\begin{equation}
	\tag{LR-I}
	\begin{aligned}
		\underset{\mathbf{q}}{min} \; & \sum\limits_{i=1}^{2k} \; q_{i} (\epsilon_{i} - e_{i})  \\
		& \textbf{s.t.} \sum\limits_{i=k+1}^{2k} c_{i} q_{i} \le \mathcal{B} \\
		& q_{i} + q_{i+k} = 1, \; i = {1,\dots,k} \\
		& 0 \le q_{i} \le 1, \; i = {1,\dots,2k}
	\end{aligned}
	\label{eq:minq2}
\end{equation}

\textbf{Observation:} The optimal solution to the linear relaxation \ref{eq:minq2} of the integer programming problem \ref{eq:minq} is integral if the cost function is uniform.

\textbf{Explanation:} The objective function of the linear relaxation can be written as
\begin{equation*}
	\begin{aligned}
		\underset{\mathbf{q}}{min} \; & \sum\limits_{i=1}^{k} \; q_{i} (\epsilon_{i} - e_{i}) \; + \; q_{i+k} (\epsilon_{i+k} - e_{i+k})\\
	\end{aligned}
\end{equation*}
Since $q_{i} + q_{i+k} = 1$, the above expression can be rewritten as
\begin{equation*}
	\begin{aligned}
		\underset{\mathbf{q}}{min} \; & \sum\limits_{i=1}^{k} \; (\epsilon_{i} - e_{i}) \; + \; q_{i+k} (\epsilon_{i+k} - e_{i+k} - (\epsilon_{i} - e_{i}))
	\end{aligned}
\end{equation*}
Therefore among the \emph{k} terms, $q_{i+k} = 1$ for the most negative values of [$(\epsilon_{i+k} - e_{i+k}) - (\epsilon_{i} - e_{i}) $] (at most $\mathcal{B}$ terms). In the remaining terms, $q_{i+k} = 0$. This explains the integrality of the optimal solution.$\qed$

Step 10 in Algorithm \ref{algo:OGDS} shows the minimization of the loss function at the $t^{th}$ iteration based on the residual error $e_{i}$ of the instance $\mathbf{x}_{i}$ given by the original classifier ($f_{\mathcal{D}}(\mathbf{x})$) and error $\epsilon_{i}$ given by the poisoned model ($ f_{{\mathcal{D}}_{p}}(\mathbf{x}) $) learned in the $(t-1)^{th}$ iteration. We iterate steps 9 to 13 until the maximum number of iterations($t_{max}$) is reached. Finally, we select the best $q$ among $t_{max}$ iterations and obtained the poisoned data $\mathcal{D}_{p}$.

\begin{table}[t]
    \centering
    \caption{Dataset Description}
    \label{tab:datasets}
    \begin{tabular}{c@{\hspace{0.5cm}} c@{\hspace{0.5cm}} c@{\hspace{0.5cm}} c}
        \toprule
        \textbf{Dataset} & \textbf{\#Instances} & \textbf{\#Features} \\ \midrule
        Ember & 12000 & 2351 \\
        Spambase & 4601 & 57 \\
        Wine & 130 & 13 \\
        Skin & 5000 & 3 \\
        Australian & 690 & 14 \\
        Banknote & 1372 & 4 \\ \bottomrule
    \end{tabular}
\end{table}

\subsection{Sorted Gradient-based Data Subversion} 
Based on the observation that the optimal solution to the linear relaxation \ref{eq:minq2} is integral, we propose sGDS (Algorithm \ref{algo:sgds}), where s in sGDS stands for \emph{``sorting"}. The choice of instances to be flipped depends on the corresponding coefficients in the objective function. Instead of solving a LP problem as in OGDS (Algorithm \ref{algo:OGDS}), this approach sorts the coefficients in each iteration (Step 10-19). To obtain the poisoned data $\mathcal{D}_{p}$, it then greedily picks the top available instances (at most $\mathcal{B}$). Despite being greedy, sGDS gives the same result as OGDS with a reduced time complexity.

\begin{algorithm}[!t]
	\KwIn{Original training data D = $\{(\mathbf{x}_{i},y_{i})\}_{i=1}^{n}$, budget B,  gradient values $\nabla = \{g_{1},\dots,g_{n}\} $, sampling ratio a and b, iteration limit $t_{max} $, adversarial cost $c_{1},\dots,c_{n}$ }
	
	$\Gamma \leftarrow$ Indices of Sort($\nabla, \textit{`desc'} )$ \\
	
	$ \nabla_{\Gamma(b)} = \nabla_{\Gamma[- b \ast \lvert D \rvert:]}$ \tcp*{Indices of top b \%}
	
	$ \nabla_{\Gamma(a)} = \nabla_{\Gamma[randA]}$ \tcp*{Indices of a \%, randA = randPick$(\{\Gamma - \Gamma(b)\},a \ast \lvert D \rvert)$}
	
	$ Ind = \nabla_{\Gamma(a)} + \nabla_{\Gamma(b)} $ \tcp*{Indices of data to be flipped, $ \lvert \nabla_{data} \rvert = $ k instances}
	
	$ f^{(0)}(\mathbf{x}) \leftarrow $ Train original classifier using data $\{(\mathbf{x}_{i}, y_{i})\}$ \\
	
	Compute \textbf{e} using $ f^{(0)}(\mathbf{x})$
	
	\bm{$\epsilon$} $ \leftarrow $ $[0]_{0}^{2k}$ ; t $ \leftarrow $ 1 \\
	
	\While{ not converge and t $\le t_{max} $}
	{
		obj $\leftarrow$ [$(\epsilon_{i} - e_{i}) $]$_{i=1}^{2k}$ \tcp*{Coefficients of objective function (refer Equation \ref{eq:minq})}
		
		obj$_\Gamma \leftarrow$ Indices of Sort(obj, $\textit{`asc'} )$ \\
		
		i $\leftarrow$ 1 ; $\#$ flips $\leftarrow$ 0 \\
		
		$\mathbf{y}^{'}(t)\leftarrow \mathbf{y}$ \\ 
		
		$choice \leftarrow [0]_{1}^{k} $ \tcp*{Indicates whether a label or its compliment is chosen}
		
		\While{i $\le$ 2k and $\#$ flips $\le$ B}
		{
			\eIf{$obj_{\Gamma_{(i)}} >$  k and $choice_{obj_{\Gamma_{(i)}} - k} $= 0 }
			{
				
				$y^{'}_{Ind({obj_{\Gamma_{(i)}} - k})} \leftarrow$  $-y_{Ind({obj_{\Gamma_{(i)}} - k})}$ \tcp*{Flip the label}
				
				$\#$ flips $\leftarrow$  $\#$ flips + 1
				
				$choice_{obj_{\Gamma_{(i)}} - k} \leftarrow $  1
			}
			{ 
				\If{$choice_{obj_{\Gamma_{(i)}}} $= 0}{
					
					$y^{'}_{Ind({obj_{\Gamma_{(i)}}})} \leftarrow$  $y_{Ind({obj_{\Gamma_{(i)}}})}$
					
					$choice_{obj_{\Gamma_{(i)}}} \leftarrow $  1
				}
			}
			
			i $\leftarrow$ i + 1
		}
		
		$f^{t}(\mathbf{x}) \leftarrow $ Retrain the classifier using the training data $ \{ (\mathbf{x}_{i},y^{'}(t))\} $ \\
		
		Compute \bm{$\epsilon$} using $f^{t}(\mathbf{x})$ \\
		
		Error (t) $\leftarrow \mathcal{V}(y_{test}, f^{(t)}(\mathbf{x}_{test}))$ \\
		
		$t \leftarrow t + 1$ \\		 
	}	
	$ t_{f} = \underset{t}{\mathbf{argmax}}$ Error \\
	
	$\mathcal{D}_{p} =  \{ (\mathbf{x}, y^{'}(t_{f}))\}_{1}^{n} $ \\				
	
	\KwOut{Poisoned training data $\mathcal{D}_{p}$ }
	
	\caption{s-Gradient-based Data Subversion (sGDS)}
	\label{algo:sgds}
\end{algorithm}

\section{Experiments and Results} \label{sec:experiments}
In this section, we validate our attack strategy on the open benchmark malware dataset EMBER (balanced set) by Endgame, Inc \cite{anderson2018ember}, spambase \cite{Dua:2019}, banknote authentication \cite{Dua:2019}, skin, wine and australian\footnote{Skin, Wine and Australian datasets can be downloaded from \url{https://www.csie.ntu.edu.tw/cjlin/libsvmtools/datasets/}} datasets. The dataset statistics is summarized in Table \ref{tab:datasets}. We evaluate the attack strategies against popular robust state-of-art classifiers: Logistic Regression (LR), LightGBM (LGBM), Naive Bayes (NB) and K-Nearest Neighbors (KNN). We used the \emph{scikit-learn}\footnote{\url{https://scikit-learn.org/stable/supervised\_learning.html}} implementation of the classifiers. The LightGBM implementation from \emph{Microsoft} \cite{ke2017lightgbm} is used for gradient ($\nabla$) computation and \emph{pyomo} with \emph{glpk solver} for the linear programming problem\footnote{\url{http://www.pyomo.org/}}. All the experiments are implemented in \emph{Python 2.7} and performed on a machine with the following configuration: 125 GB RAM, 96 CPUs, Model: Intel (R) Xeon (R) Platinum 8160 CPU @ 2.10GHz.

\begin{figure}[t]
	\centering 
	\begin{subfigure}{0.33\textwidth}
		\includegraphics[width=\linewidth]{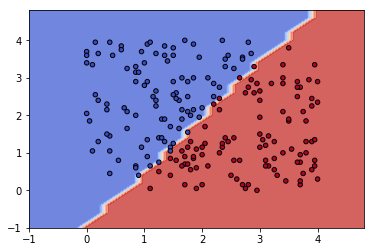}
		\caption{Original\\ classifier}
		\label{fig:linearOriginal}
	\end{subfigure}\hfil 
	\begin{subfigure}{0.33\textwidth}
		\includegraphics[width=\linewidth]{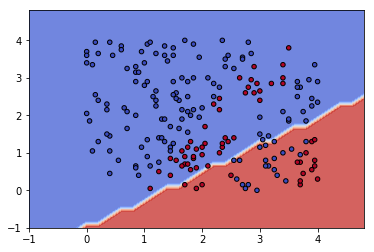}
		\caption{Poisoned\\ classifier}
		\label{fig:linearPoisoned}
	\end{subfigure}\hfil 
	\begin{subfigure}{0.33\textwidth}
		\includegraphics[width=\linewidth]{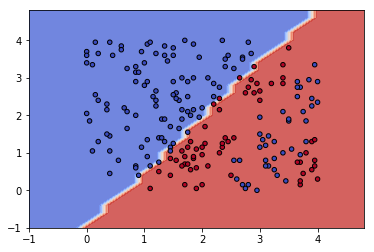}
		\caption{Poisoned data on original\\ classifier}
		\label{fig:linearPoisonedOriginal}
	\end{subfigure}
	\medskip
	\begin{subfigure}{0.33\textwidth}
		\includegraphics[width=\linewidth]{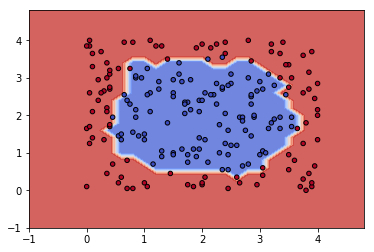}
		\caption{Original\\ classifier}
		\label{fig:circOriginal}
	\end{subfigure}\hfil 
	\begin{subfigure}{0.33\textwidth}
		\includegraphics[width=\linewidth]{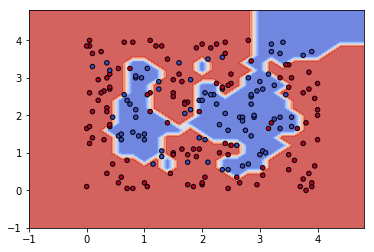}
		\caption{Poisoned\\ classifier}
		\label{fig:circPoisoned}
	\end{subfigure}\hfil 
	\begin{subfigure}{0.33\textwidth}
		\includegraphics[width=\linewidth]{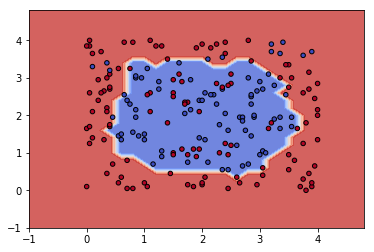}
		\caption{Poisoned data on original\\ classifier}
		\label{fig:circPoisnedOriginal}
	\end{subfigure}
	\caption{Visualizing the GDS attack in synthetic datasets (linear and circular) for two learning algorithms: Logistic Regression (LR) and K-Nearest Neighbor (KNN). The data instances are represented as points and based on the labels instances are colored as red and blue circles.}
	\label{fig:VisualizeAttacks}
\end{figure}

\subsection{\bf Result Analysis}
To evaluate GDS and OGDS (Algorithm \ref{algo:GDS}, \ref{algo:OGDS}), we compare the classification errors with two baselines. The first baseline is the Linear flip strategy (Linear), where the attacker flips a fraction of small gradient data instances of the training data for a given budget ($\mathcal{B}$). The second baseline is the Projected Gradient Ascent (PGA) algorithm (dual logistic regression) \cite{zhao2017efficient}. We demonstrate the performance of GDS and OGDS with four different training classifiers under various adversarial budgets. The values of \emph{a} and \emph{b} for our strategies are chosen based on the empirical analysis of different combinations which together ([\emph{a+b}]$\times$ 100\%) do not constitute more than 50-70\% of the training data. 

\begin{table}[t]
    \centering
    \caption{Performance comparison of approaches with Logistic Regression as classifier in multiple datasets.}
    \label{tab:resultUCI}
    \begin{tabular}{|l|c|ll|ll|ll|}
    \hline
    \multirow{2}{*}{\textbf{Dataset}} & \multirow{2}{*}{\textbf{Actual Error}} & \multicolumn{2}{c}{\textbf{10\%}} & \multicolumn{2}{c}{\textbf{20\%}} & \multicolumn{2}{c|}{\textbf{30\%}} \\ \cline{3-8} 
     &  & \multicolumn{1}{c@{\hspace{0.5cm}}}{PGA} & \multicolumn{1}{c}{s-GDS} & \multicolumn{1}{c@{\hspace{0.5cm}}}{PGA} & \multicolumn{1}{c}{s-GDS} & \multicolumn{1}{c@{\hspace{0.5cm}}}{PGA} & \multicolumn{1}{c|}{s-GDS} \\ \hline
    Spambase & 0.068 & 0.167 & \bf{0.193} & 0.239 & \bf{0.319} & 0.377 & \bf{0.455} \\
    Wine & 0.015 & \bf{0.123} & \bf{0.123} & 0.277 & \bf{0.292} & 0.423 & \bf{0.503} \\
    Skin & 0.083 & \bf{0.234} & 0.156 & 0.229 & \bf{0.318} & 0.392 & \bf{0.446} \\
    Banknote & 0.019 & 0.093 & \bf{0.105} & 0.202 & \bf{0.223} & 0.320 & \bf{0.535} \\
    Ember & 0.411 & 0.471 & \bf{0.521} & 0.455 & \bf{0.53} & 0.526 & \bf{0.539} \\
    Australian & 0.122 & 0.161 & \bf{0.197} & 0.197 & \bf{0.339} & 0.306 & \bf{0.545} \\ \hline
    \end{tabular}
\end{table}

We utilize two-dimensional synthetic data to visualize the degradation in the model under OGDS. We generated the dataset with linear and circular pattern. We have 1000 instances, out of which 200 instances are selected for training and 800 for testing purpose. The attack visualization is for 20\% label contamination. Figure \ref{fig:VisualizeAttacks} illustrates, 
\begin{enumerate}
	\item \emph{Original classifier}: decision boundary of model trained on the clean data points (Figure \ref{fig:linearOriginal}, \ref{fig:circOriginal}),
	\item \emph{Poisoned classifier}: decision boundary of model trained on the poisoned data points under OGDS (Figure \ref{fig:linearPoisoned}, \ref{fig:circPoisoned}), and
	\item \emph{Poisoned data on original classifier}: the position of contaminated labels in the original classifier (Figure \ref{fig:linearPoisonedOriginal}, \ref{fig:circPoisnedOriginal});
\end{enumerate}
In both linear and circular synthetic data sets, we can observe the changes in the decision boundary under our attack. The actual error rate and error rate of the classifiers (LR for linear data and K-NN for the circular data) under attack is tabulated in Table \ref{tab:resultsyn}.

\begin{table}[ht]
    \centering
    \caption{Performance of our approach in synthetic datasets under various budgets. The tuned parameter (a,b) is (0.01, 0.49).}
    \label{tab:resultsyn}
    \begin{tabular}{@{}lc@{\hspace{0.5cm}} c@{\hspace{0.5cm}} c@{\hspace{0.5cm}} c@{}}
    \toprule
    \textbf{Dataset} & \multicolumn{1}{l}{\textbf{Actual Error}} & \multicolumn{1}{l}{\textbf{10\%}} & \multicolumn{1}{l}{\textbf{20\%}} & \multicolumn{1}{l}{\textbf{30\%}} \\ \midrule
    Linear (LR) & 0.045 & 0.191 & 0.356 & 0.53 \\
    Circular (K-NN) & 0.054 & 0.150 & 0.27 & 0.399 \\ \bottomrule
    \end{tabular}
\end{table}

\begin{table}[!t]
    \centering
    \caption{Transferability of attack using different surrogate models (with subscript $s$). Error rate of victim model (with subscript $v$) is shown in the table. Each cell denotes the error rate of the victim model under attack for corresponding surrogate model. To evaluate the transferability for surrogate models, we have highlighted the maximum error-rate other than diagonal values.}
    \label{tab:transferUCI}
    \begin{tabular}{c@{\hspace{0.6cm}} c@{\hspace{0.6cm}} c@{\hspace{0.6cm}} c @{\hspace{0.6cm}} c @{\hspace{0.6cm}} c}
    \toprule
     & \multicolumn{5}{c}{\textbf{Wine Dataset}} \\ \midrule
     & LGBM$_{v}$ & KNN$_{v}$ & NB$_{v}$ & LR$_{v}$ & SVM$_{v}$ \\
     LGBM$_{s}$ & 0.292 & 0.269 & 0.038 & 0.246 & 0.3 \\
     KNN$_{s}$ & 0.292 & 0.446 & \bf{0.162} & \bf{0.362} & \bf{0.308} \\
     NB$_{s}$ & \bf{0.3} & 0.254 & 0.123 & 0.285 & 0.3 \\
     LR$_{s}$ & \bf{0.3} & \bf{0.423} & 0.1 & 0.508 & 0.3 \\
     SVM$_{s}$ & 0.285 & 0.223 & 0.015 & 0.138 & 0.3 \\ \midrule
     & \multicolumn{5}{c}{\textbf{Australian Dataset}} \\ \midrule
     LGBM$_{s}$ & 0.3 & 0.325 & 0.175 & \bf{0.325} & 0.314 \\
     KNN$_{s}$ & \bf{0.301} & 0.425 & \bf{0.467} & 0.216 & 0.362 \\
     NB$_{s}$ & 0.299 & 0.4 & 0.522 & 0.241 & 0.368 \\
     LR$_{s}$ & \bf{0.301} & 0.433 & 0.242 & 0.551 & \bf{0.483} \\
     SVM$_{s}$ & 0.3 & \bf{0.442} & 0.291 & 0.232 & 0.422 \\ \midrule
    & \multicolumn{5}{c}{\textbf{Spambase Dataset}} \\ \midrule
    LGBM$_{s}$ & 0.29 & 0.352 & 0.196 & \bf{0.389} & 0.339 \\
    KNN$_{s}$ & 0.295 & 0.426 & 0.351 & 0.271 & 0.332 \\
    NB$_{s}$ & 0.283 & 0.328 & 0.423 & 0.303 & 0.277 \\
    LR$_{s}$ & \bf{0.314} & 0.363 & \bf{0.625} & 0.458 & 0.351 \\
    SVM$_{s}$ & 0.295 & \bf{0.456} & 0.394 & 0.288 & \bf{0.412} \\ \midrule
   & \multicolumn{5}{c}{\textbf{Banknote Dataset}} \\ \midrule
   LGBM$_{s}$ & 0.298 & \bf{0.328} & 0.324 & 0.396 & \bf{0.35} \\
   KNN$_{s}$ & \bf{0.302} & 0.305 & 0.296 & 0.157 & 0.281 \\
   NB$_{s}$ & 0.3 & 0.3 & 0.592 & \bf{0.407} & 0.297 \\
   LR$_{s}$ & 0.3 & 0.313 & \bf{0.534} & 0.532 & 0.319 \\
   SVM$_{s}$ & 0.3 & 0.327 & 0.324 & 0.402 & 0.351 \\ \midrule
  & \multicolumn{5}{c}{\textbf{Skin Dataset}} \\ \midrule
  LGBM$_{s}$ & 0.301 & 0.296 & 0.214 & 0.224 & 0.299 \\
  KNN$_{s}$ & \bf{0.327} & 0.304 & 0.512 & 0.437 & \bf{0.311} \\
  NB$_{s}$ & 0.311 & \bf{0.313} & 0.521 & \bf{0.438} & 0.31 \\
  LR$_{s}$ & 0.316 & 0.311 & \bf{0.513} & 0.439 & 0.31 \\
  SVM$_{s}$ & 0.205 & 0.244 & 0.172 & 0.076 & 0.294 \\ \bottomrule
    \end{tabular}
    \end{table}

The error rate of the classifiers is compared and reported in Table \ref{tab:resultUCI}. We can observe that the s-GDS attack is performing better with Wine, Skin, Banknote and Australian dataset and achieves around 44-54\% error rate by flipping 30\% of the data points. Even though, our approach has better error rate on EMBER dataset, we can see that Logistic Regression is a weak classifier for the dataset (having poor performance in clean dataset). Moreover, we can see that PGA performs better with Spambase dataset. Moreover, in some dataset (wine and skin with 10\% budget) PGA performs better and almost similar as our approach. The prior work \cite{koh2018stronger} posed the question: \emph{``What conditions make certain models on certain datasets attackable, but not others?"} and speculated that the higher dimensionality and linearly inseparable nature of data helped to attack the linear models on those datasets. We postulate that the higher dimensionality of spambase nature of dataset results in the variation of error rate between PGA and s-GDS.

\subsubsection*{\bf Transferability of Attack} In the real world scenario, an attacker does not have any knowledge of the victim model. Even though she does not have any information about the architecture of the victim learning model, she can perform the attack which is designed for other learning model \cite{zhao2017efficient}. 

In order to evaluate our attack strategy, we analysed the transferability of attack with multiple datasets (similar to \cite{zhao2017efficient}). We considered Logistic Regression (LR), naive bayes (NB), lightGBM (LGBM), SVM, and K-Nearest Neighbors (KNN) as both surrogate and victim models. Also, the attacker's budget is set as 30\% of the training set size. Table \ref{tab:transferUCI} shows the error rates of the victim models under the OGDS attack, which was computed against the aforementioned surrogate models. We did not compare transferability with Demontis et al. \cite{transferUsenix} as their focus is on studying the factors that contribute to the transferability of attacks, therefore it is not most suitable for comparison. Since Zhao et al. \cite{zhao2017efficient} provides the baseline attack for our work, we compared transferability against it and found that our OGDS attack has comparable transferability, for example, OGDS has good transferability when LR is used as surrogate, whereas \cite{zhao2017efficient} has better transferability with SVM as surrogate. When kNN is victim model then OGDS has better transferability, whereas for NB, \cite{zhao2017efficient} performs better. For DT and LightGBM, the error rates are similar for both of us. 

\begin{figure}[!t] 
	\begin{subfigure}[b]{0.24\textwidth}
		\centering
		\includegraphics[width=1\linewidth]{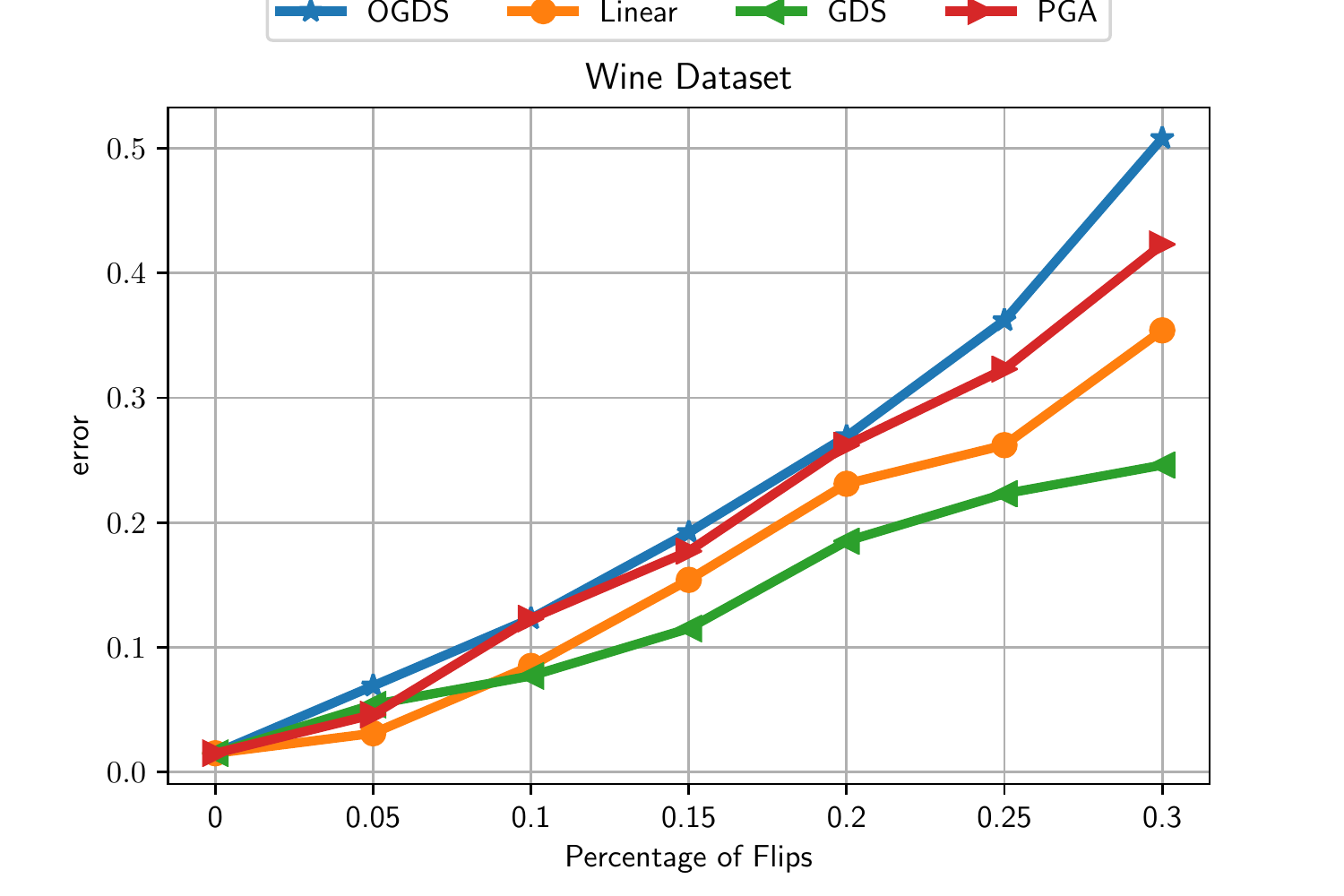} 
		\caption{LR} 
		\label{fig:LRwine} 
	\end{subfigure}
	\begin{subfigure}[b]{0.24\textwidth}
		\centering
		\includegraphics[width=1\linewidth]{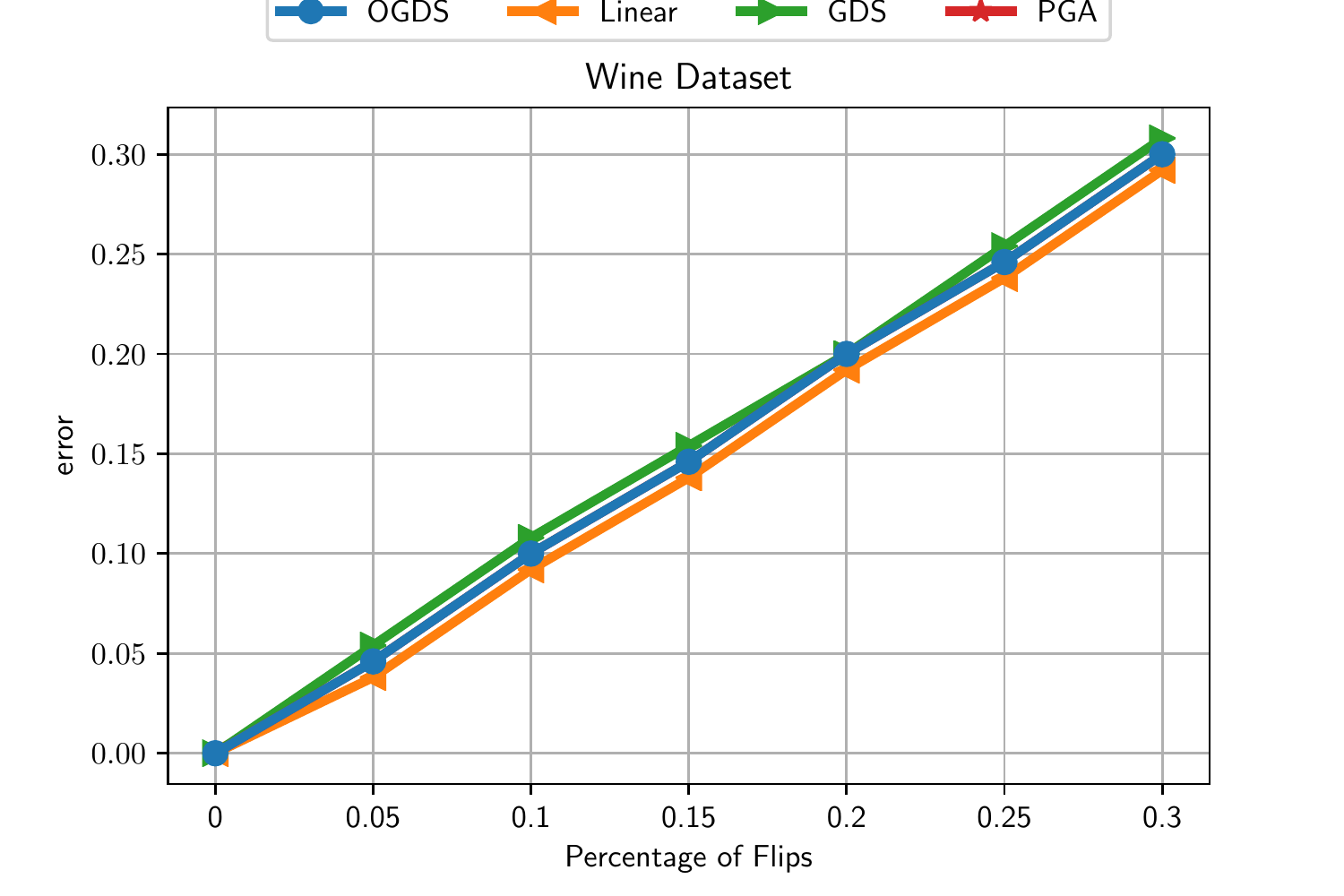} 
		\caption{LightGBM}
		\label{fig:lgbmWine} 
	\end{subfigure}
	\begin{subfigure}[b]{0.24\textwidth}
		\centering
		\includegraphics[width=1\linewidth]{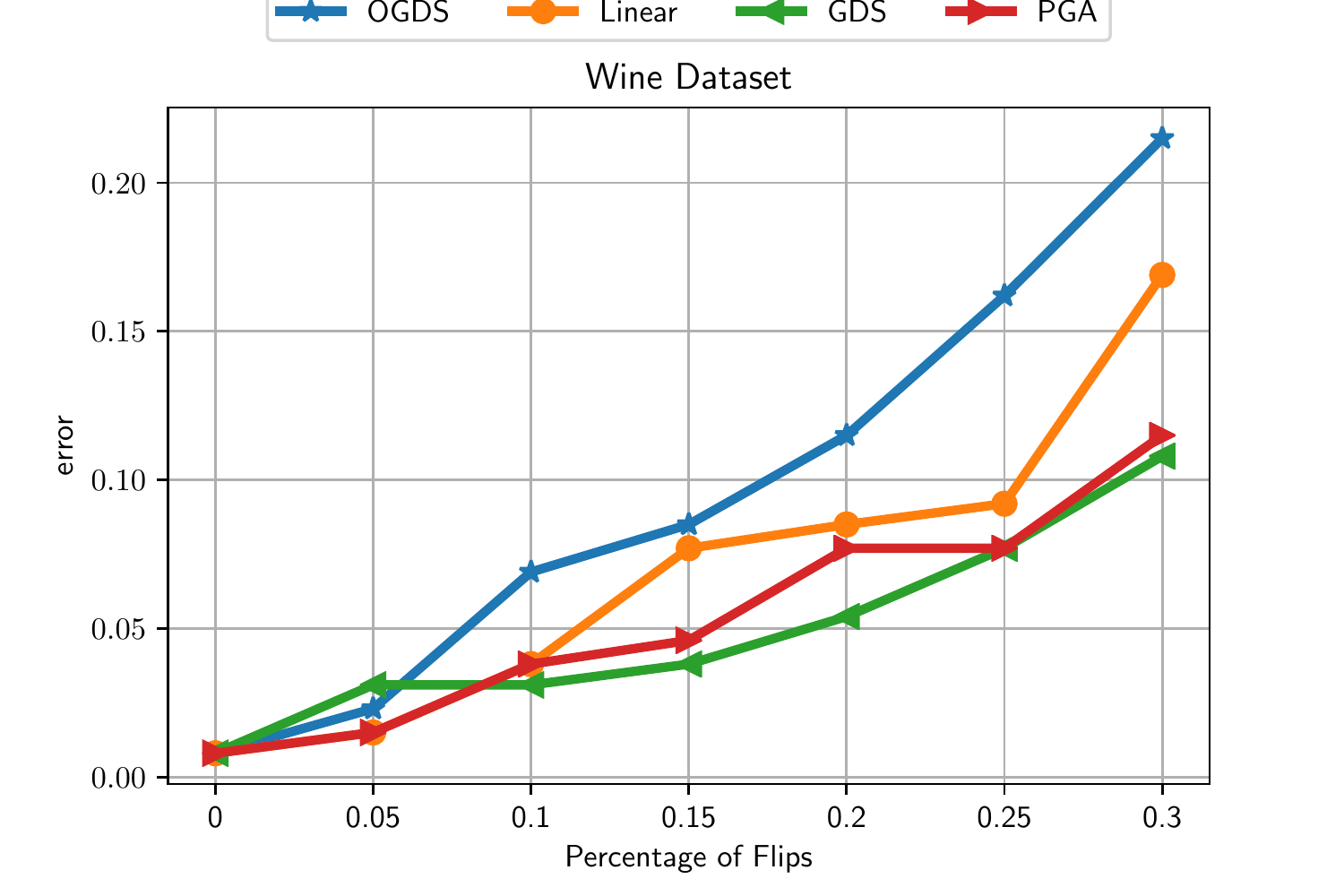} 
		\caption{Naive Bayes}
		\label{fig:NBwine}
	\end{subfigure}
	\begin{subfigure}[b]{0.24\textwidth}
		\centering
		\includegraphics[width=1\linewidth]{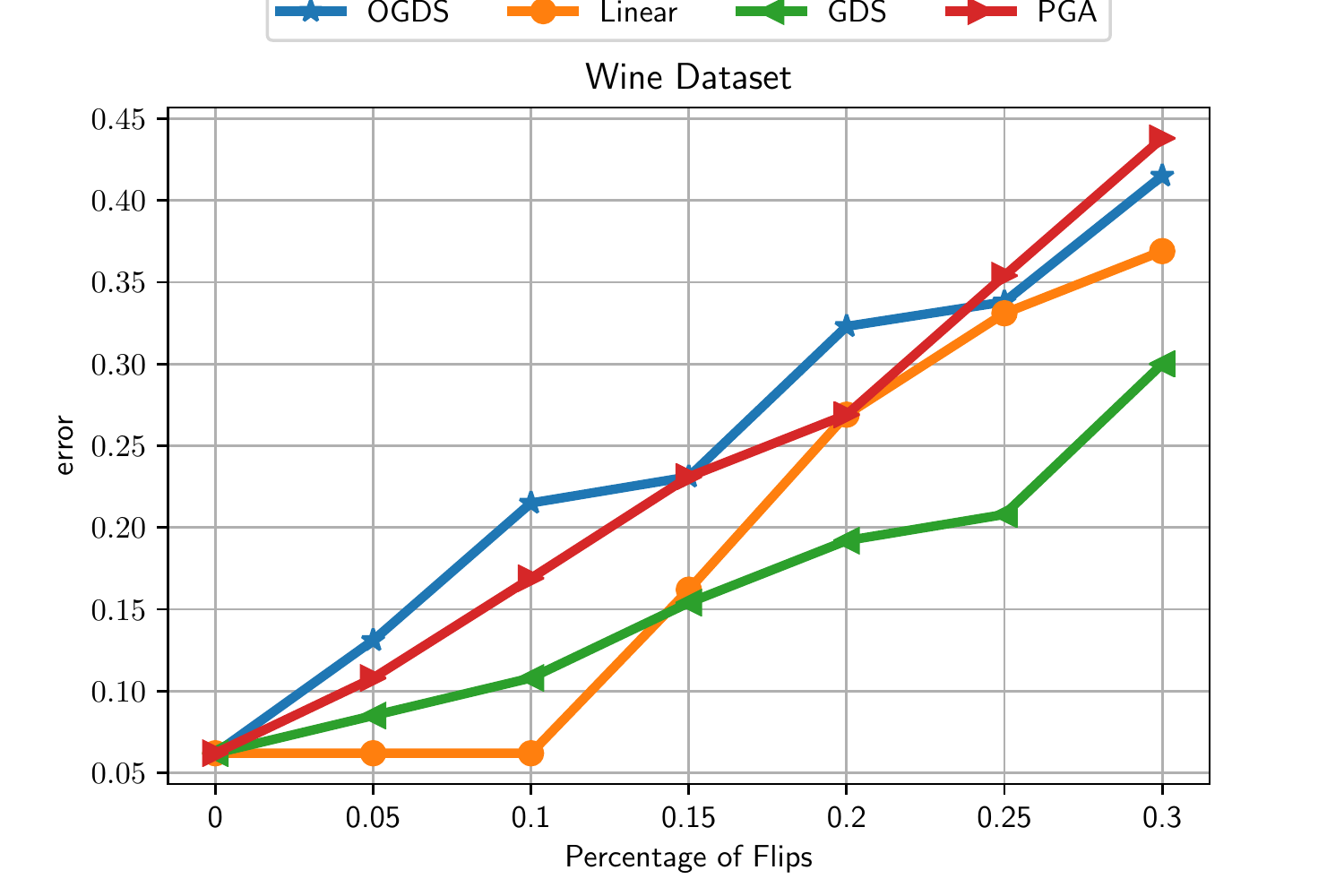} 
		\caption{KNN} 
		\label{fig:knnWine} 
	\end{subfigure} 
	
	\begin{subfigure}[b]{0.24\textwidth}
		\centering
		\includegraphics[width=1\linewidth]{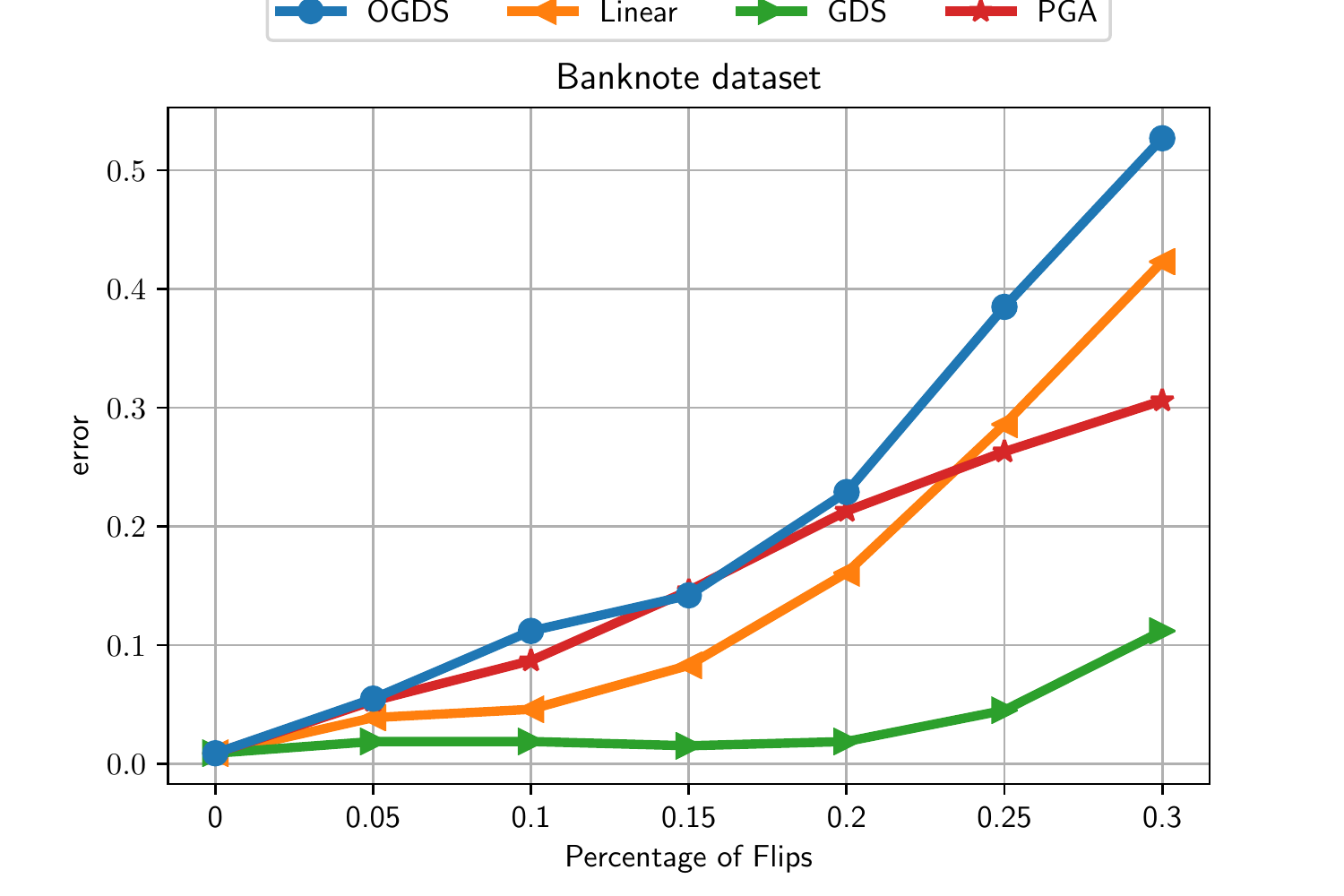} 
		\caption{LR}  
		\label{fig:LRbank} 
	\end{subfigure} 
	\begin{subfigure}[b]{0.24\textwidth}
		\centering
		\includegraphics[width=1\linewidth]{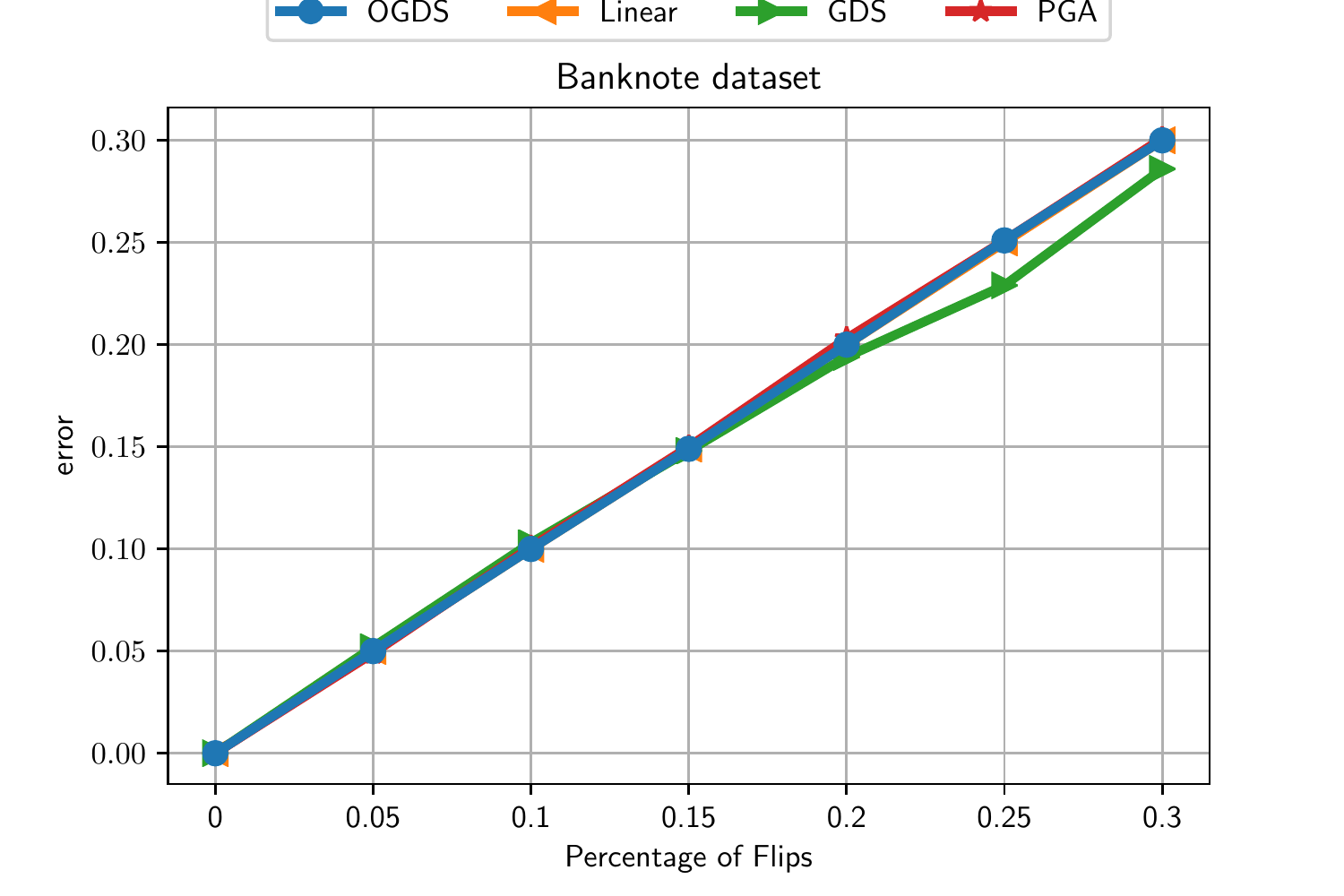} 
		\caption{LightGBM}
		\label{fig:lgbmbank}
	\end{subfigure}
	\begin{subfigure}[b]{0.24\textwidth}
		\centering
		\includegraphics[width=1\linewidth]{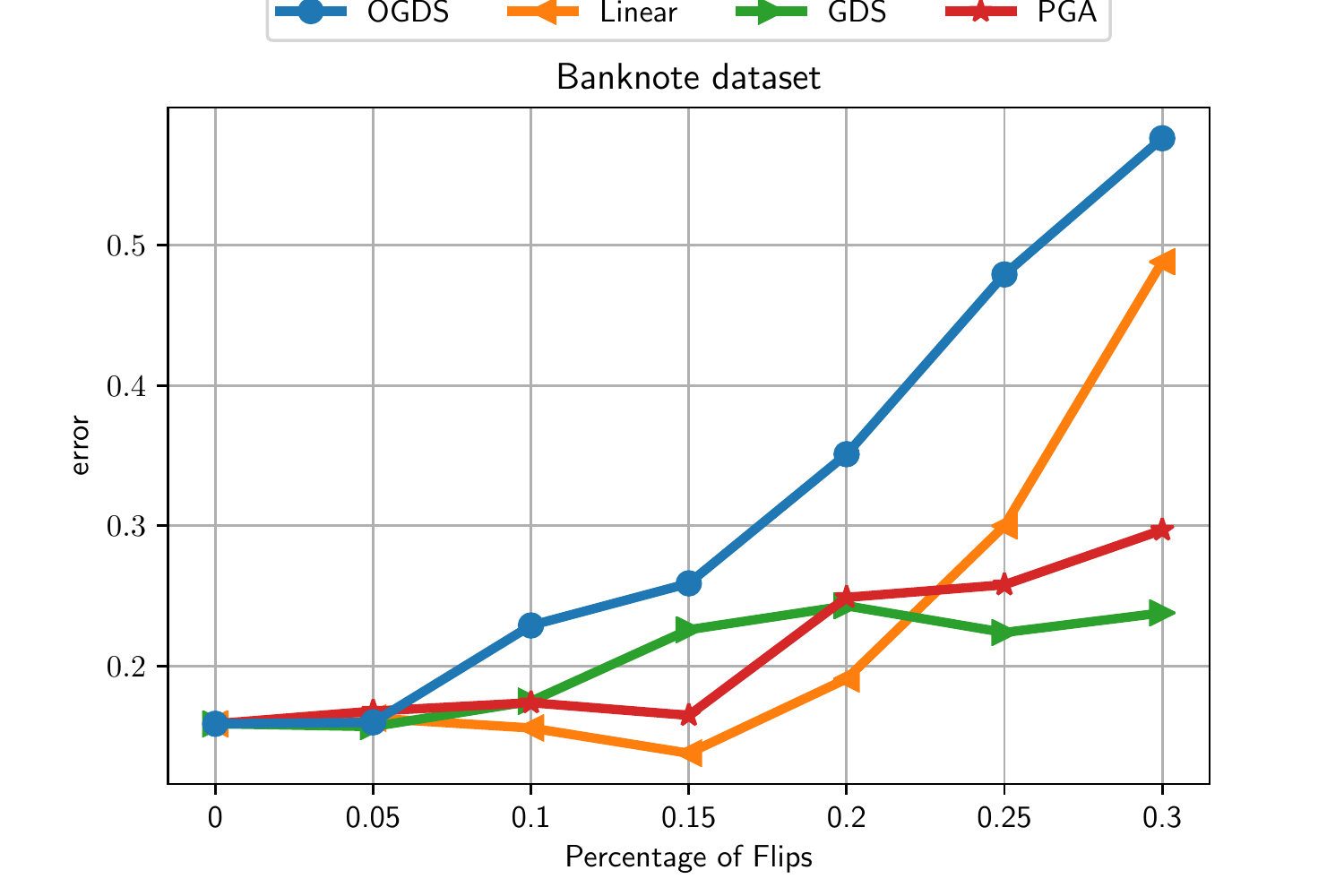} 
		\caption{Naive Bayes}
		\label{fig:NBbank}
	\end{subfigure} 
	\begin{subfigure}[b]{0.24\textwidth}
		\centering
		\includegraphics[width=1\linewidth]{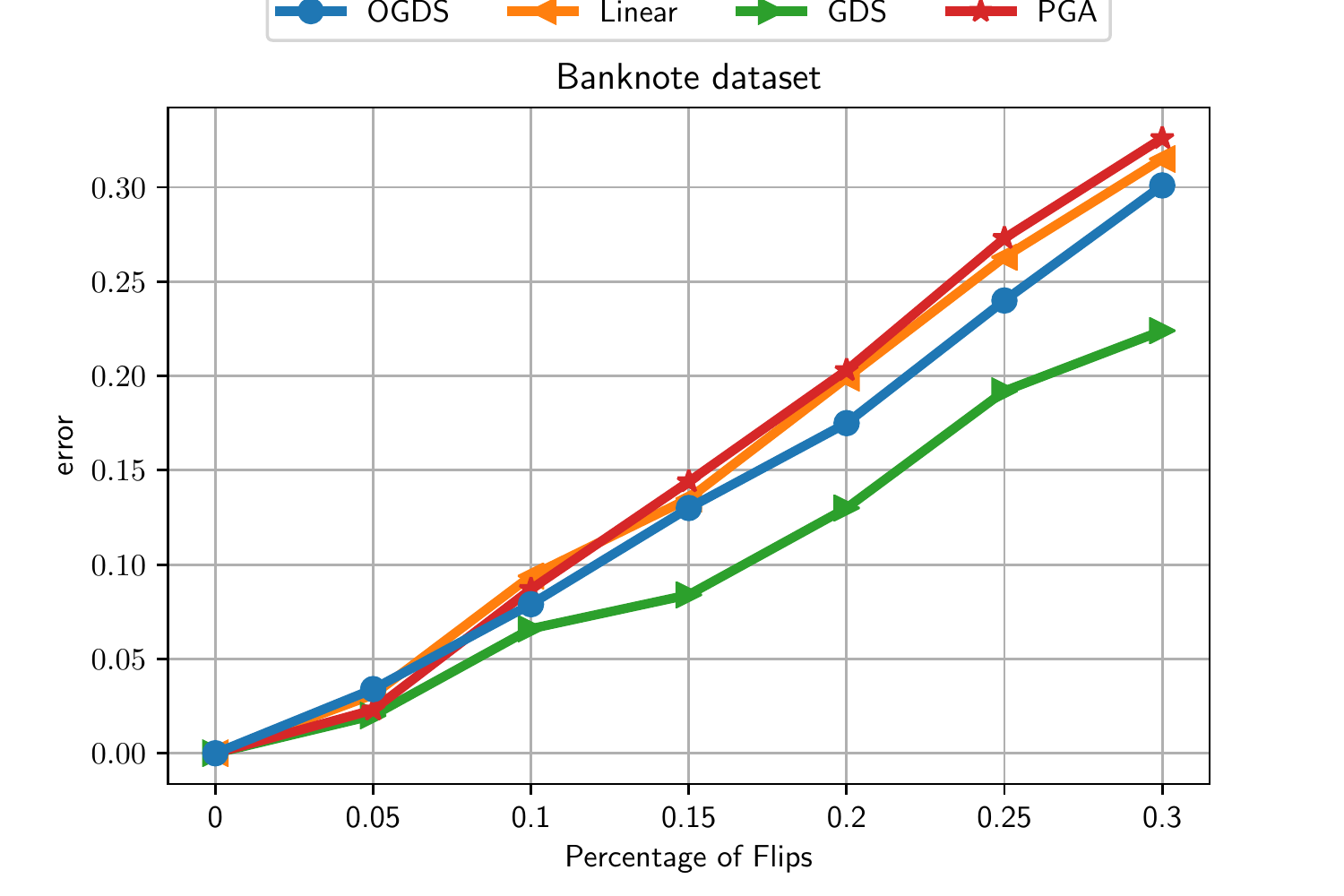} 
		\caption{KNN}  
		\label{fig:knnbank}
	\end{subfigure}
	
	\begin{subfigure}[b]{0.24\textwidth}
		\centering
		\includegraphics[width=1\linewidth]{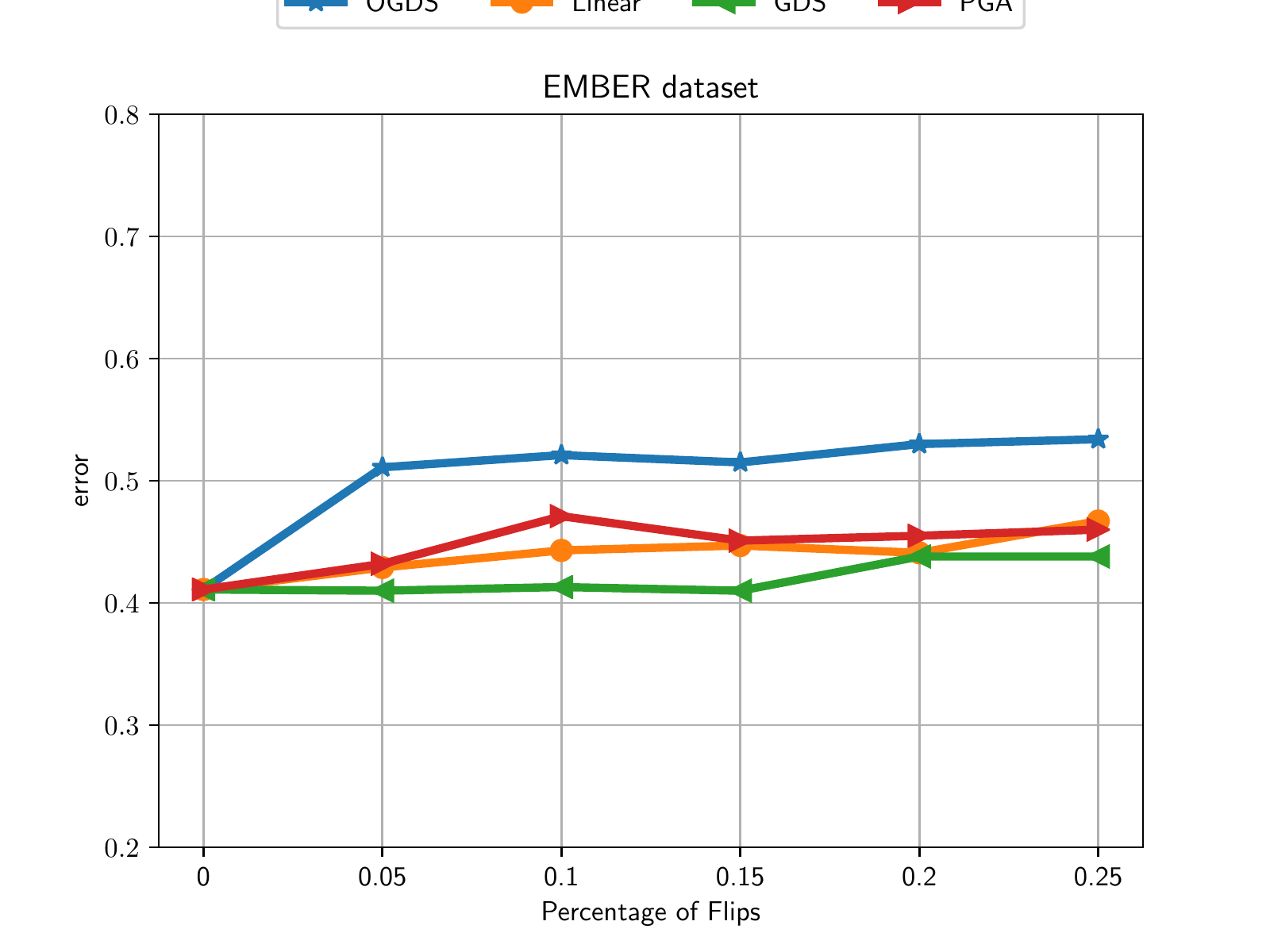} 
		\caption{LR}  
		\label{fig:LRember}
	\end{subfigure}
	\begin{subfigure}[b]{0.24\textwidth}
		\centering
		\includegraphics[width=1\linewidth]{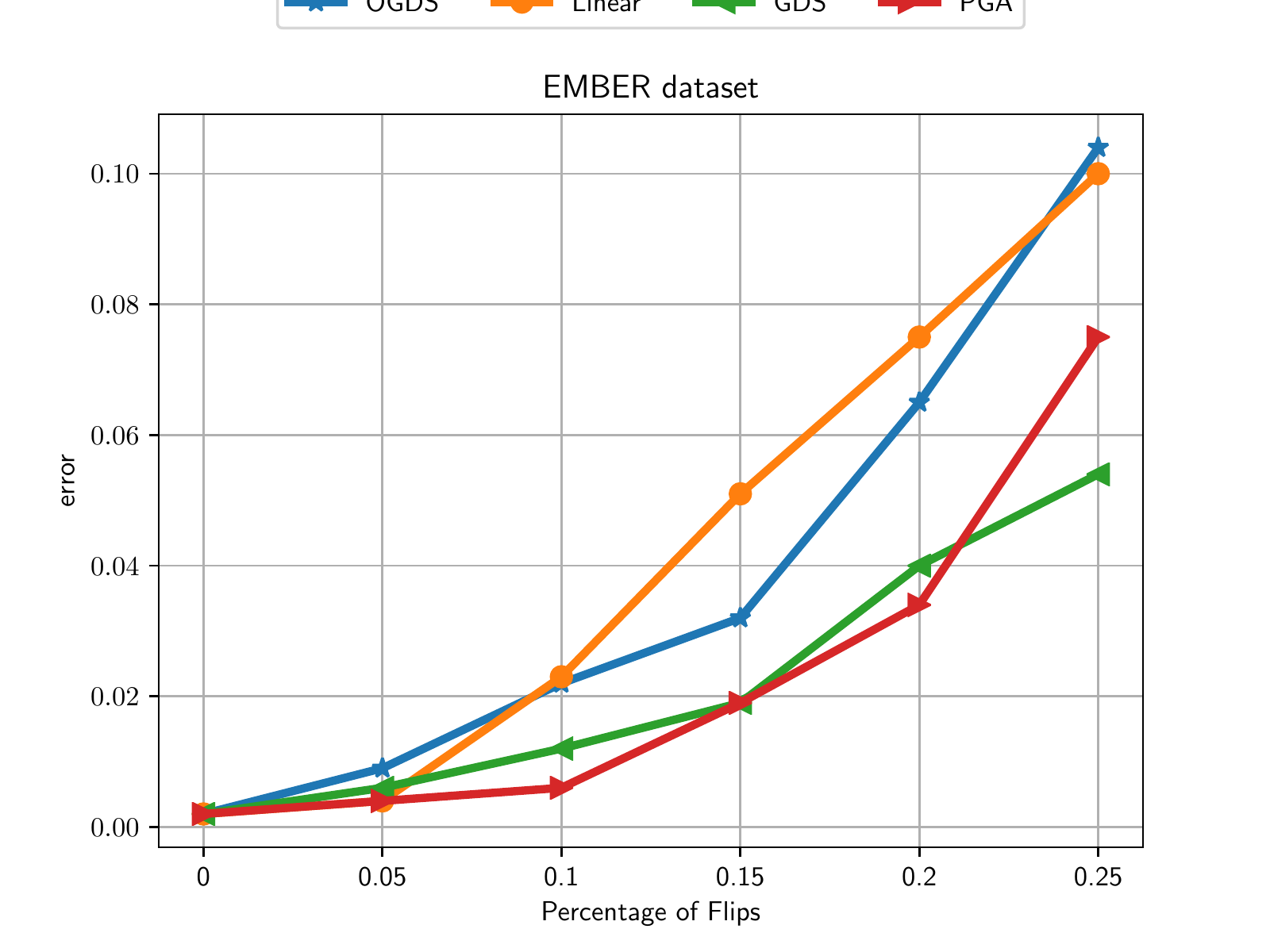} 
		\caption{LightGBM}
		\label{fig:lgbmEmber} 
	\end{subfigure}
	\begin{subfigure}[b]{0.24\textwidth}
		\centering
		\includegraphics[width=1\linewidth]{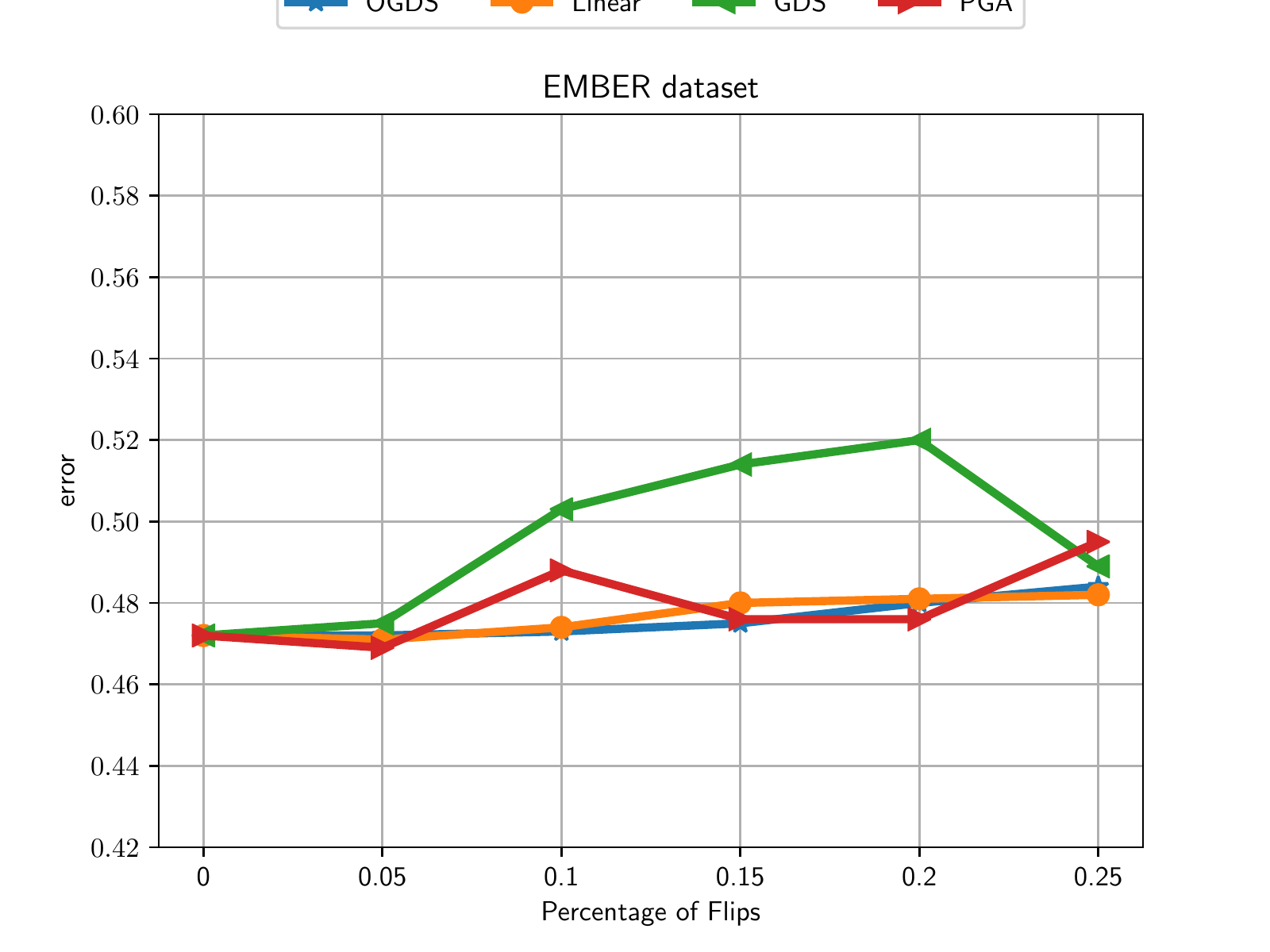} 
		\caption{Naive Bayes}
		\label{fig:NBember}
	\end{subfigure}
	\begin{subfigure}[b]{0.24\textwidth}
		\centering
		\includegraphics[width=1\linewidth]{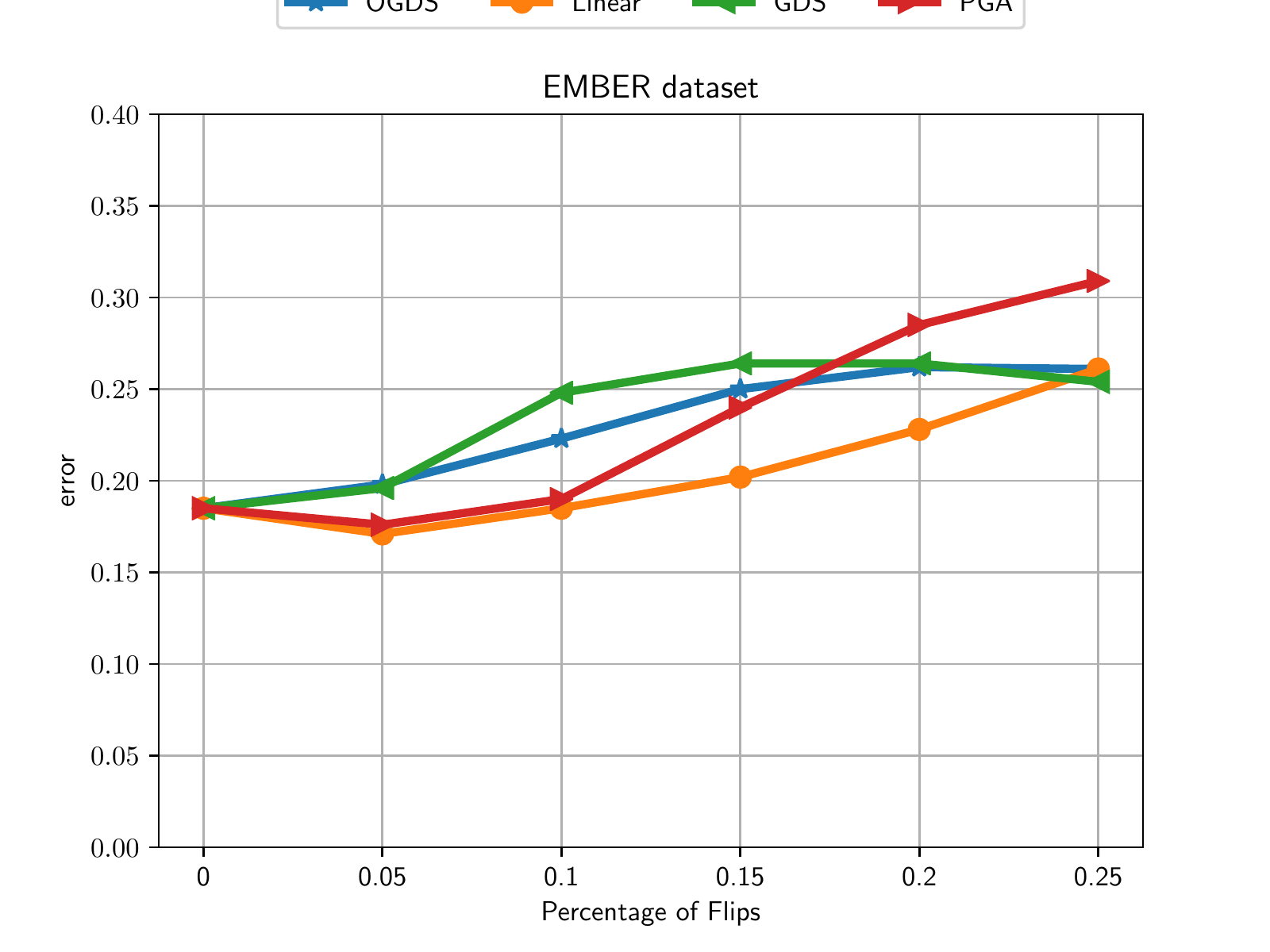} 
		\caption{KNN} 
		\label{fig:knnember}
	\end{subfigure} 
	
	\begin{subfigure}[b]{0.24\textwidth}
		\centering
		\includegraphics[width=1\linewidth]{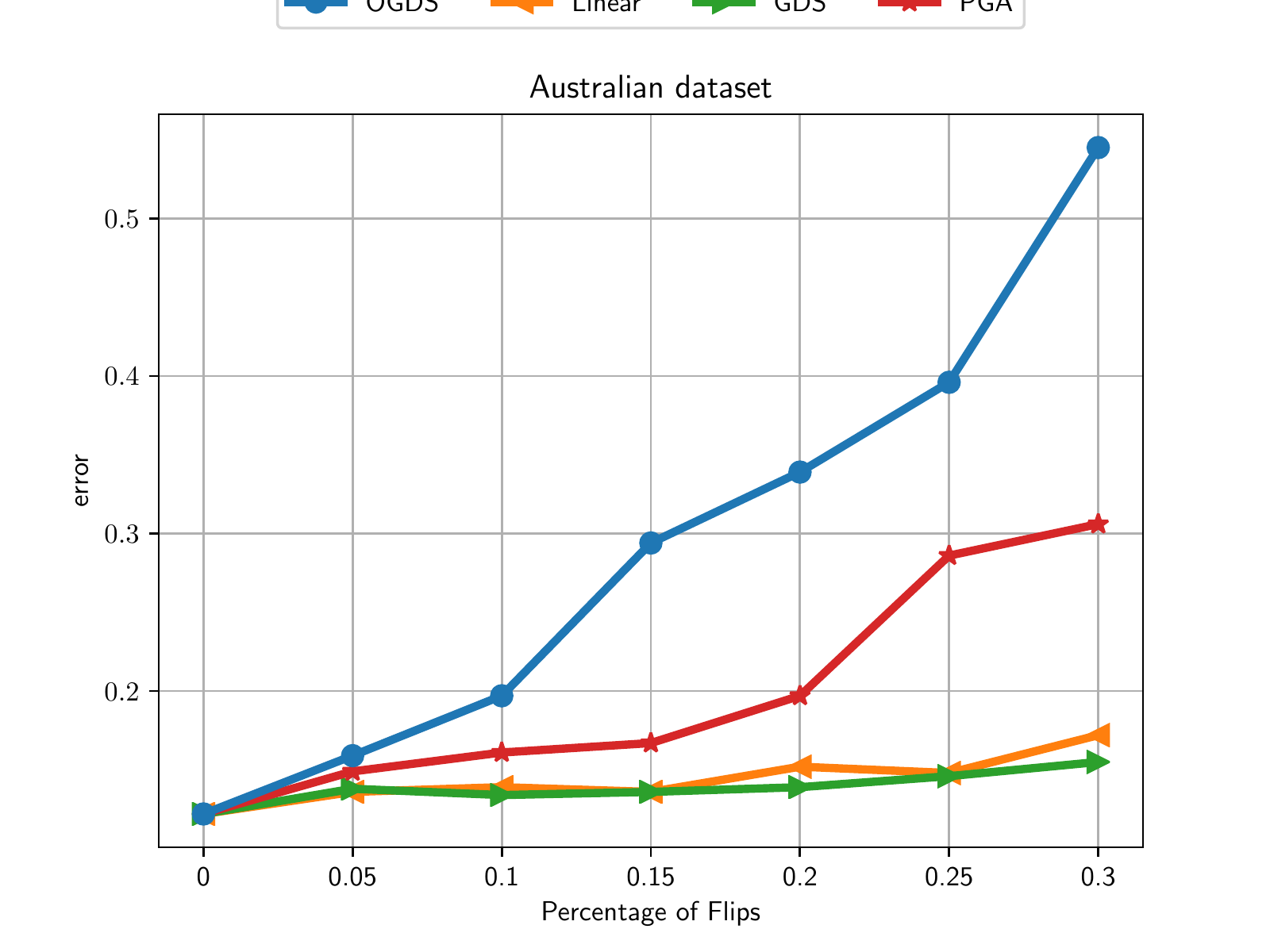} 
		\caption{LR}  
		\label{fig:LRaus}
	\end{subfigure}
	\begin{subfigure}[b]{0.24\textwidth}
		\centering
		\includegraphics[width=1\linewidth]{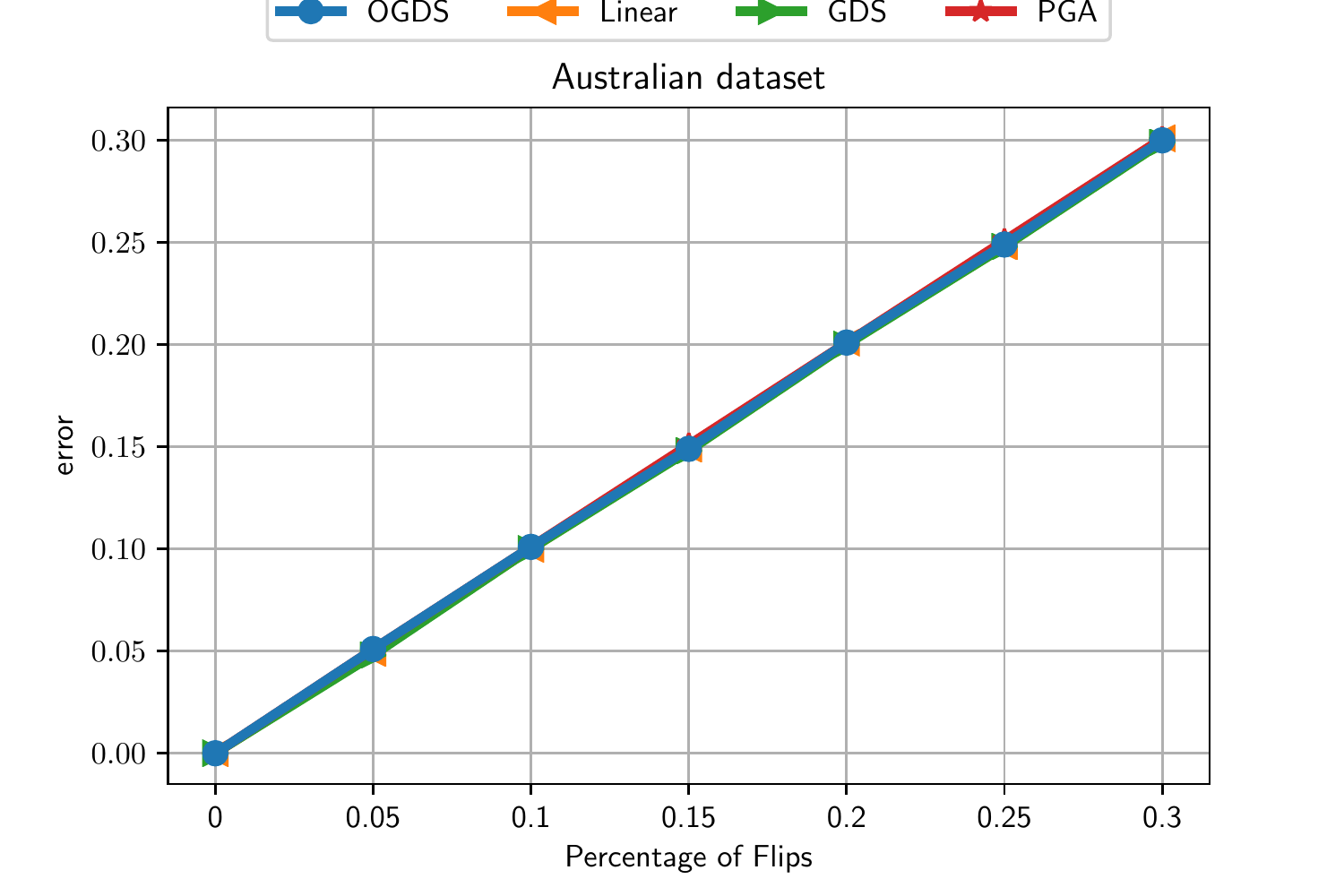} 
		\caption{LightGBM}  
		\label{fig:LGBMaus}
	\end{subfigure}
	\begin{subfigure}[b]{0.24\textwidth}
		\centering
		\includegraphics[width=1\linewidth]{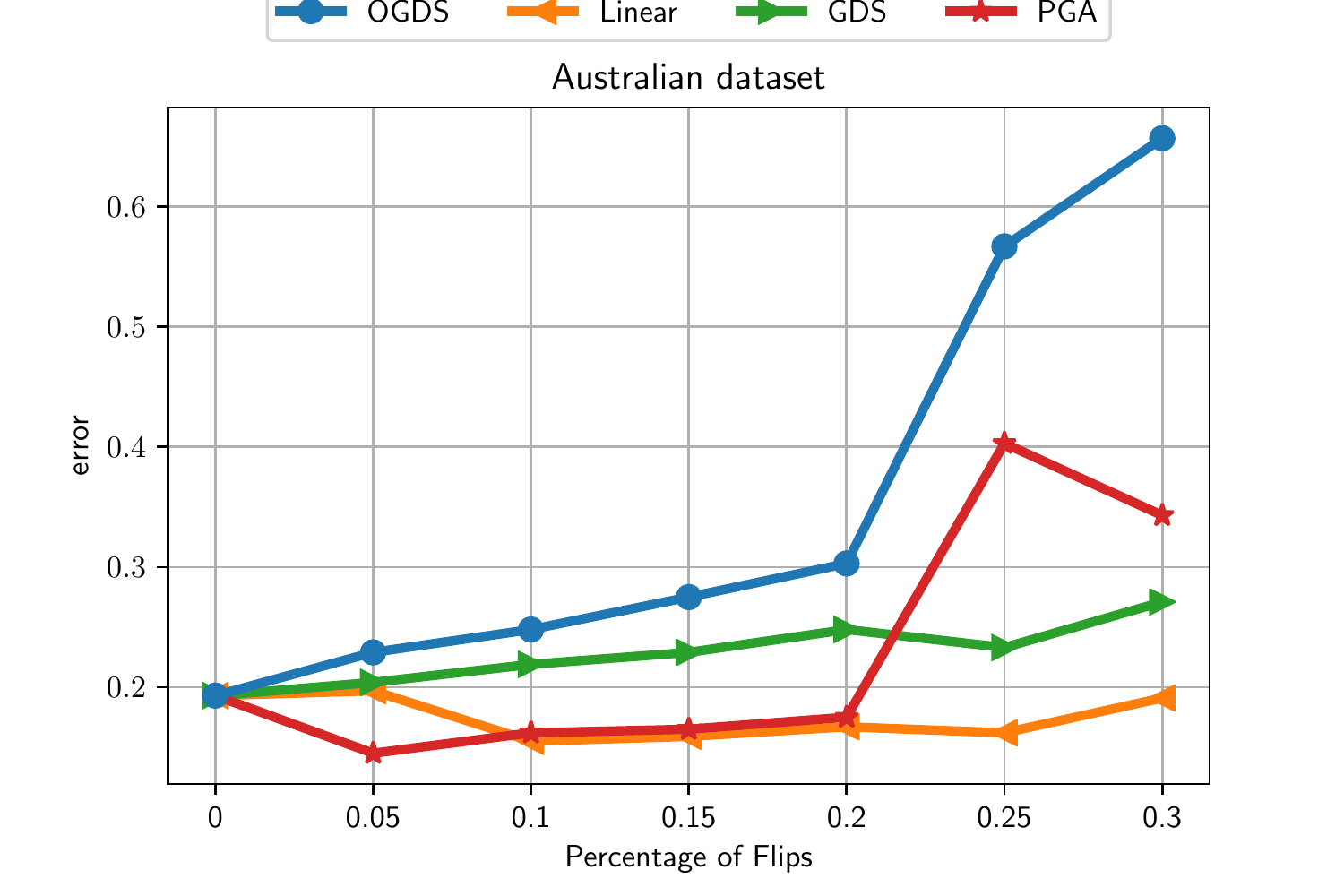} 
		\caption{Naive Bayes}  
		\label{fig:NBaus}
	\end{subfigure}
	\begin{subfigure}[b]{0.24\textwidth}
		\centering
		\includegraphics[width=1\linewidth]{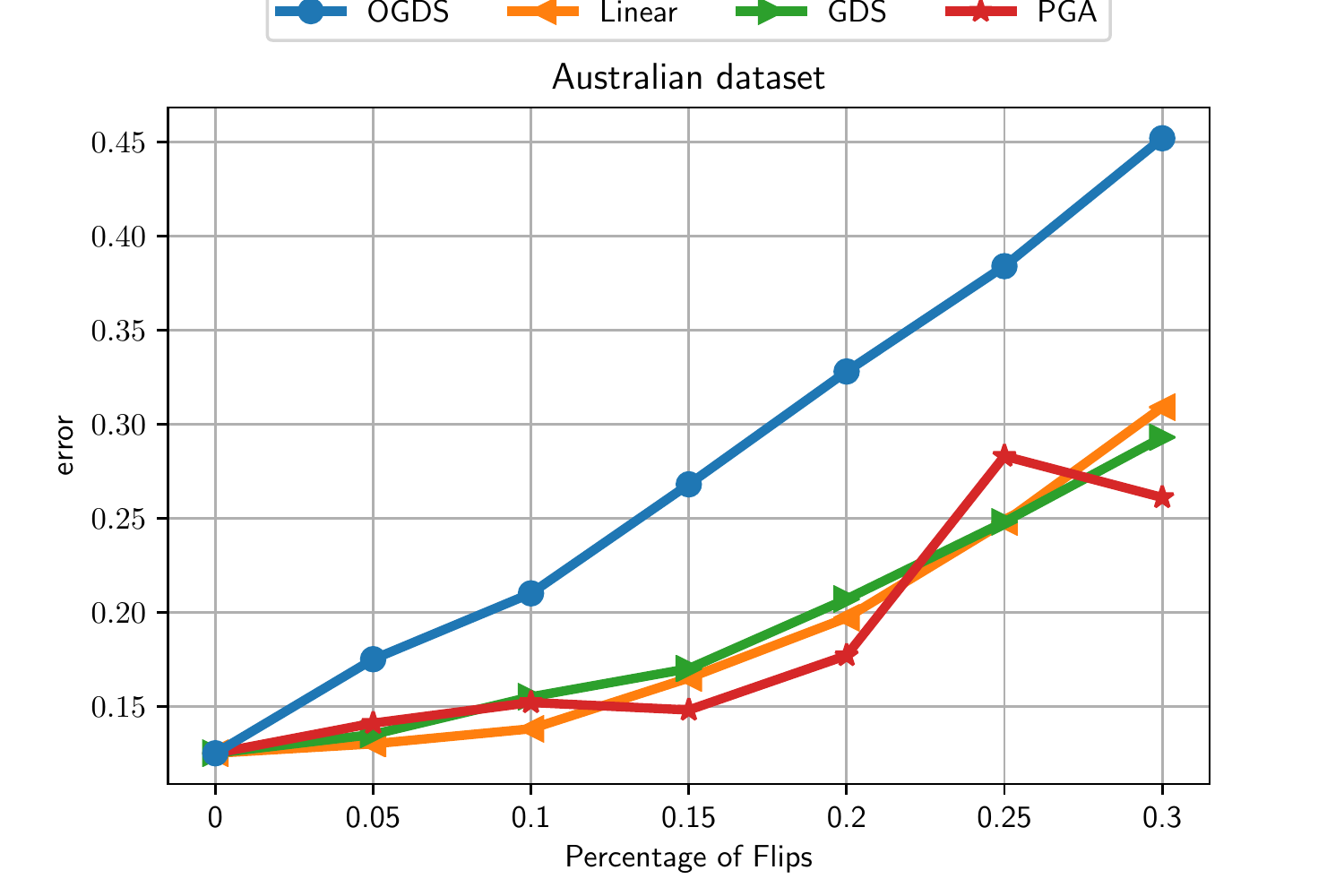} 
		\caption{KNN}  
		\label{fig:knnaus}
	\end{subfigure}
	
	\begin{subfigure}[b]{0.24\textwidth}
		\centering
		\includegraphics[width=1\linewidth]{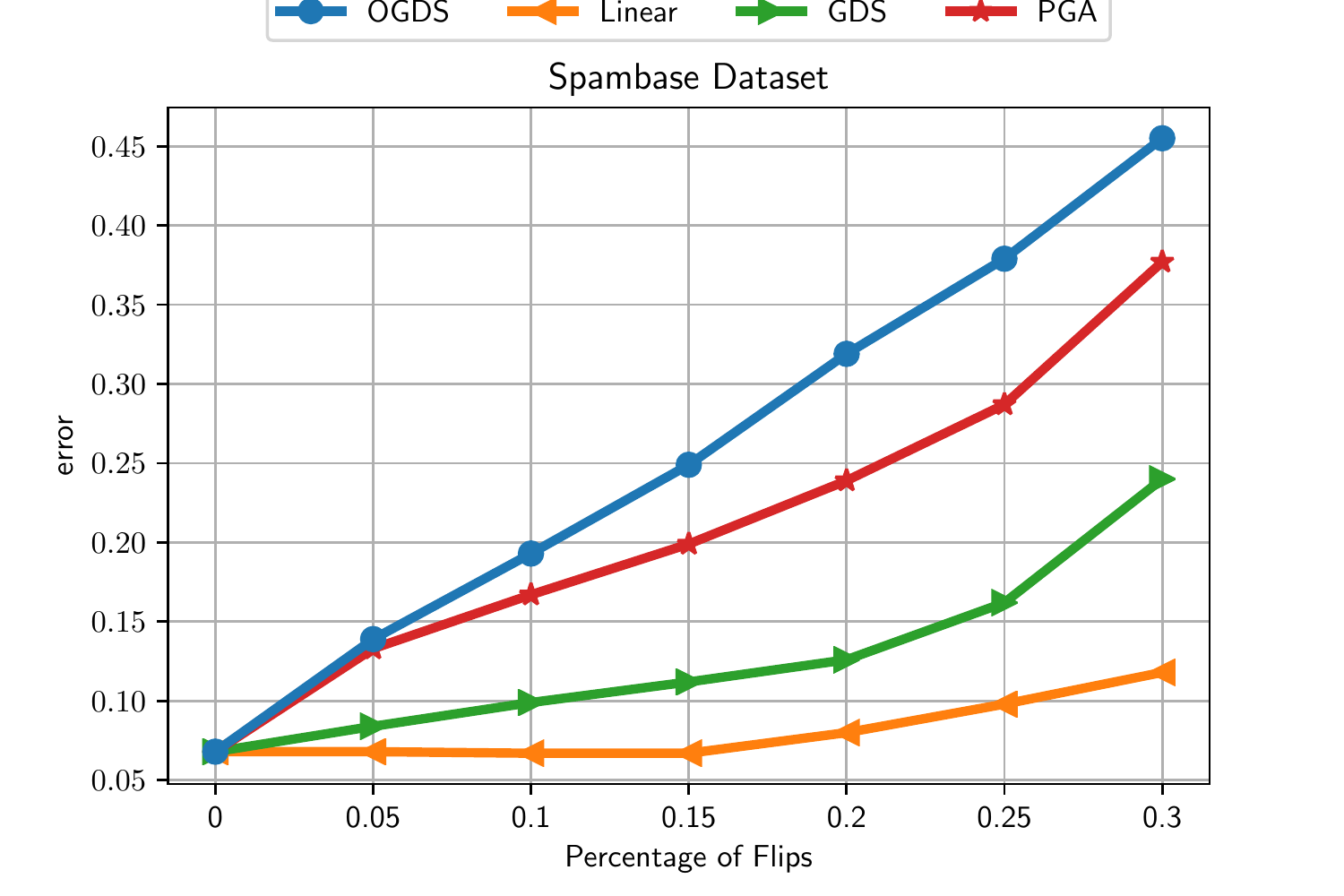} 
		\caption{LR}  
		\label{fig:LRspam}
	\end{subfigure}
	\begin{subfigure}[b]{0.24\textwidth}
		\centering
		\includegraphics[width=1\linewidth]{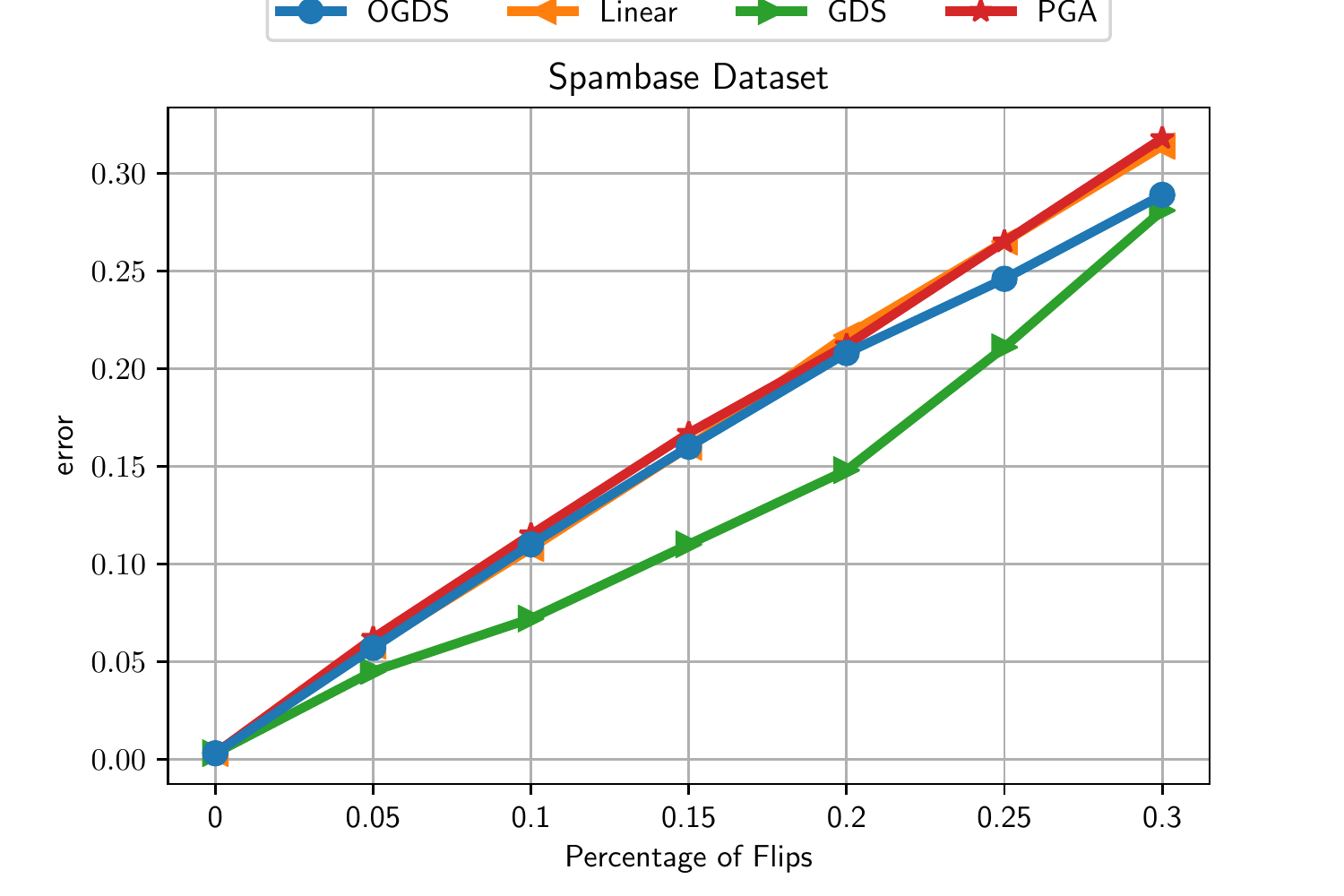} 
		\caption{LightGBM}  
		\label{fig:Lgbmspam}
	\end{subfigure}
	\begin{subfigure}[b]{0.24\textwidth}
		\centering
		\includegraphics[width=1\linewidth]{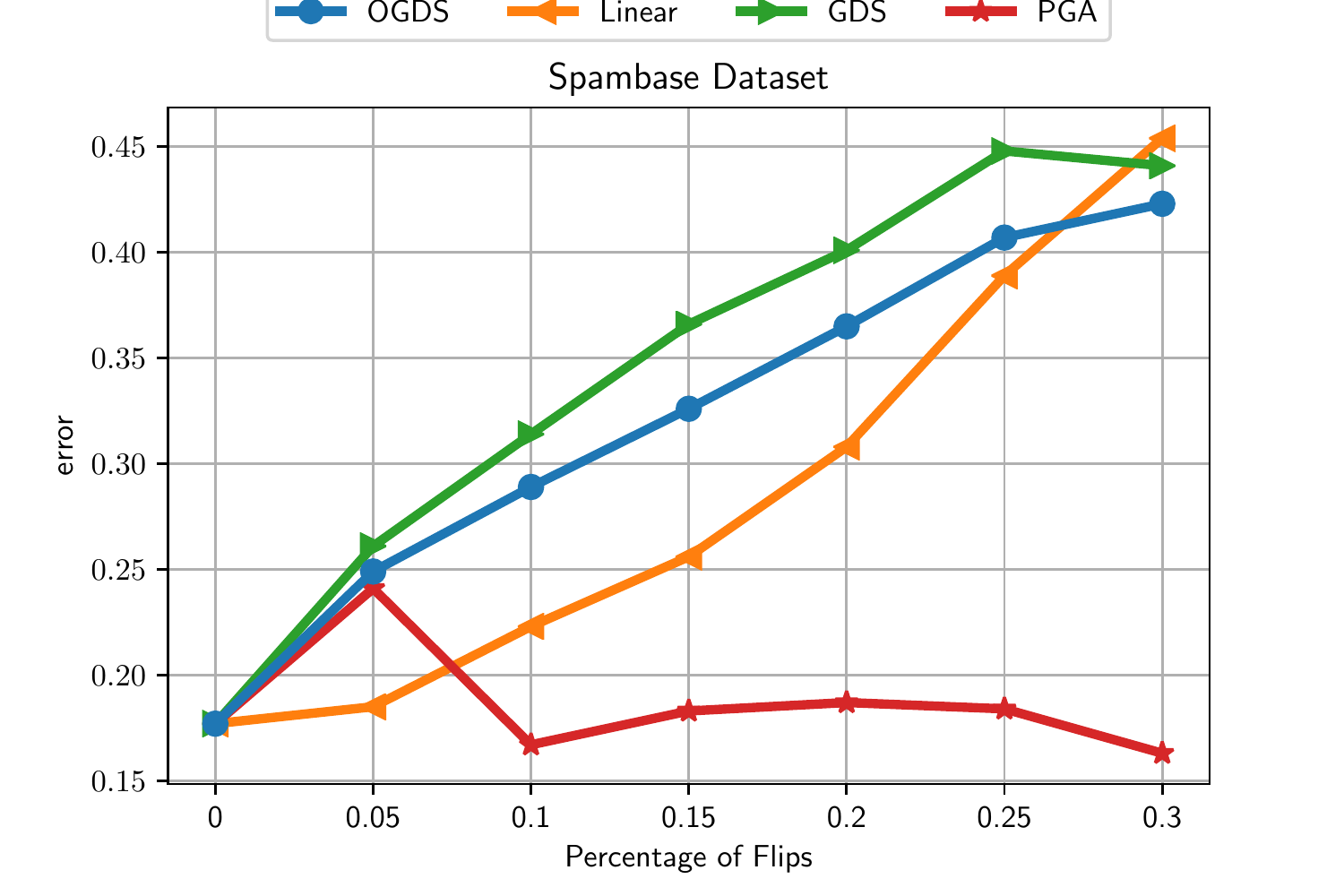} 
		\caption{Naive Bayes}  
		\label{fig:NBspam}
	\end{subfigure}
	\begin{subfigure}[b]{0.24\textwidth}
		\centering
		\includegraphics[width=1\linewidth]{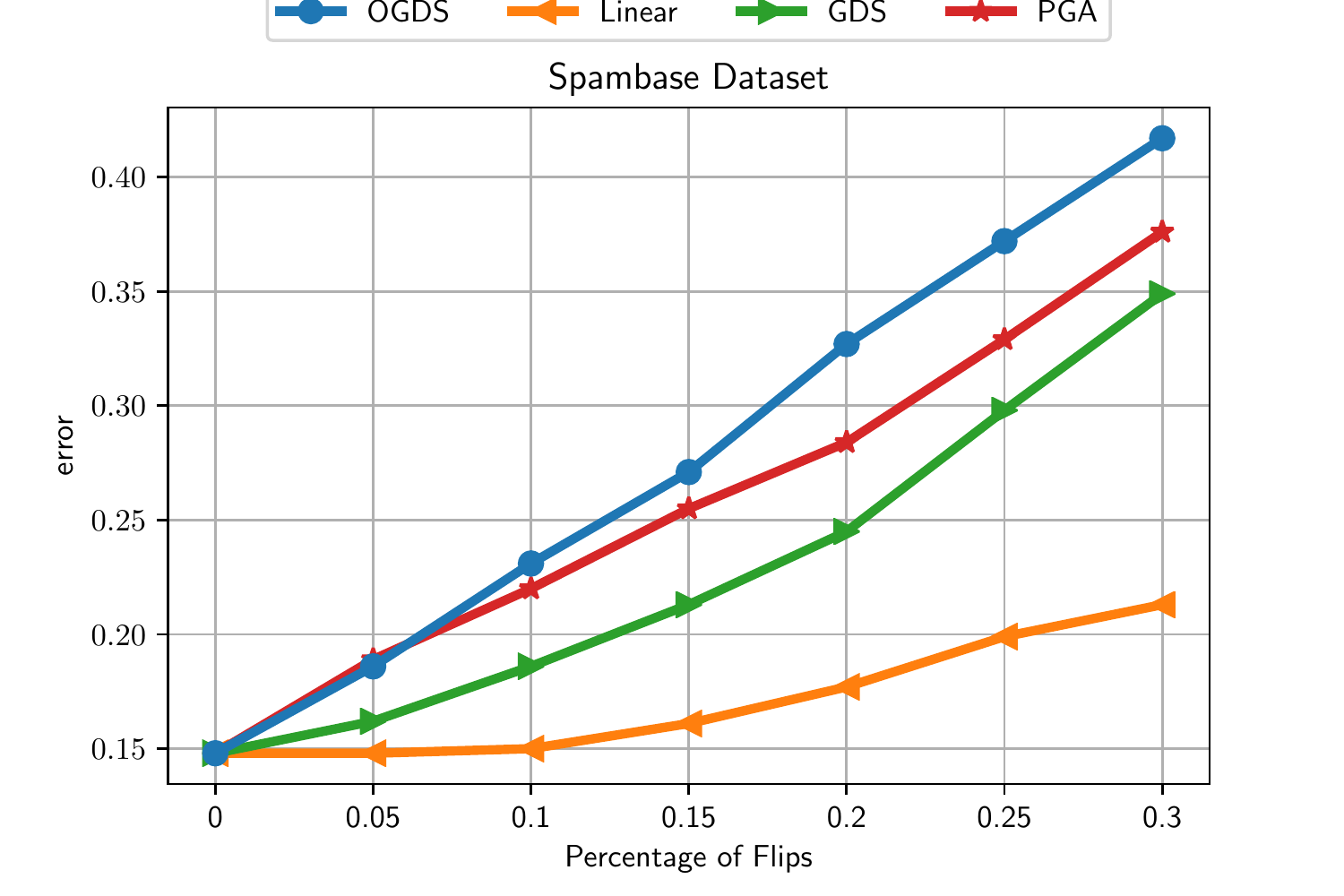} 
		\caption{KNN}  
		\label{fig:knnspam}
	\end{subfigure}
	
	\begin{subfigure}[b]{0.24\textwidth}
		\centering
		\includegraphics[width=1\linewidth]{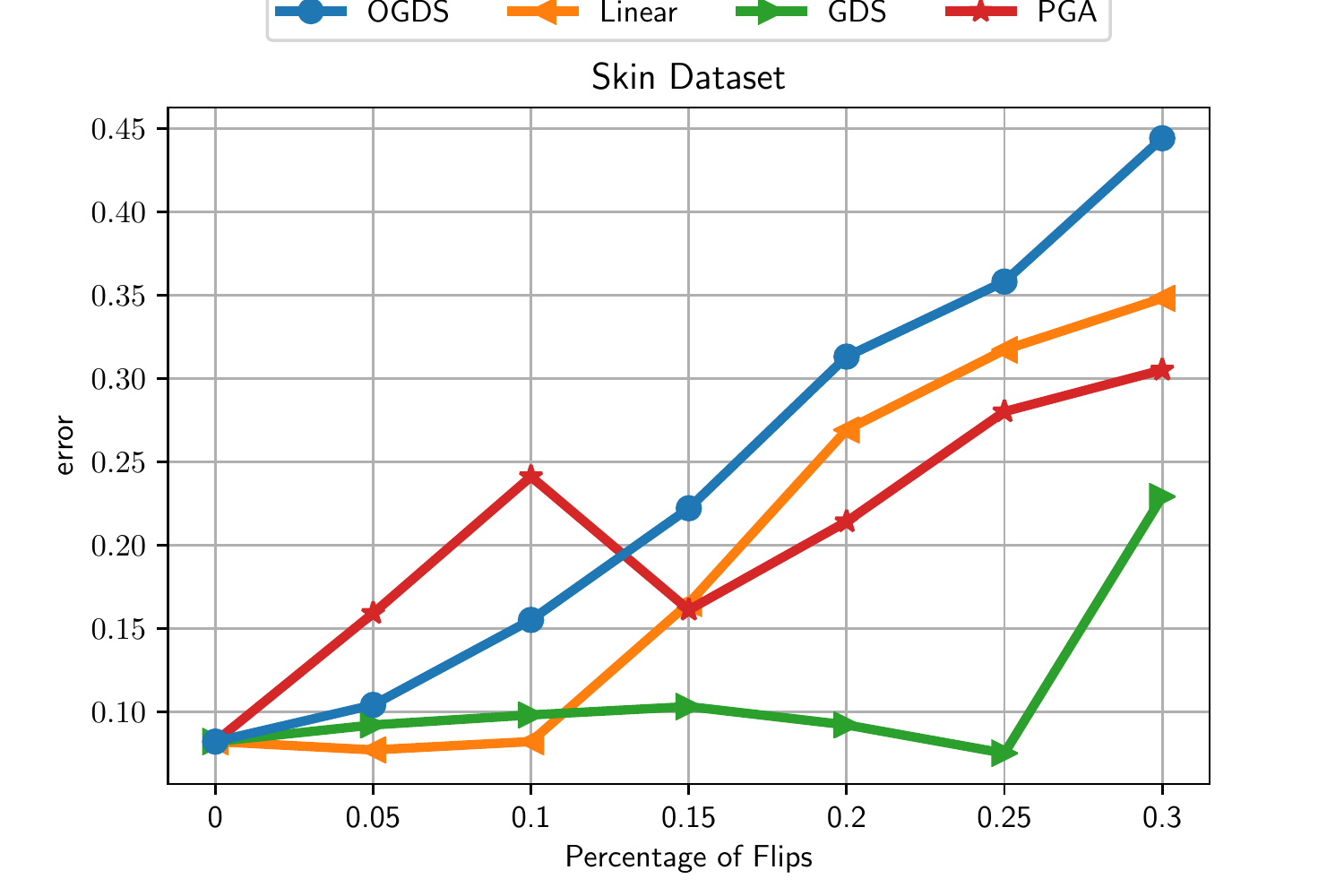} 
		\caption{LR}  
		\label{fig:LRskin}
	\end{subfigure}
	\begin{subfigure}[b]{0.24\textwidth}
		\centering
		\includegraphics[width=1\linewidth]{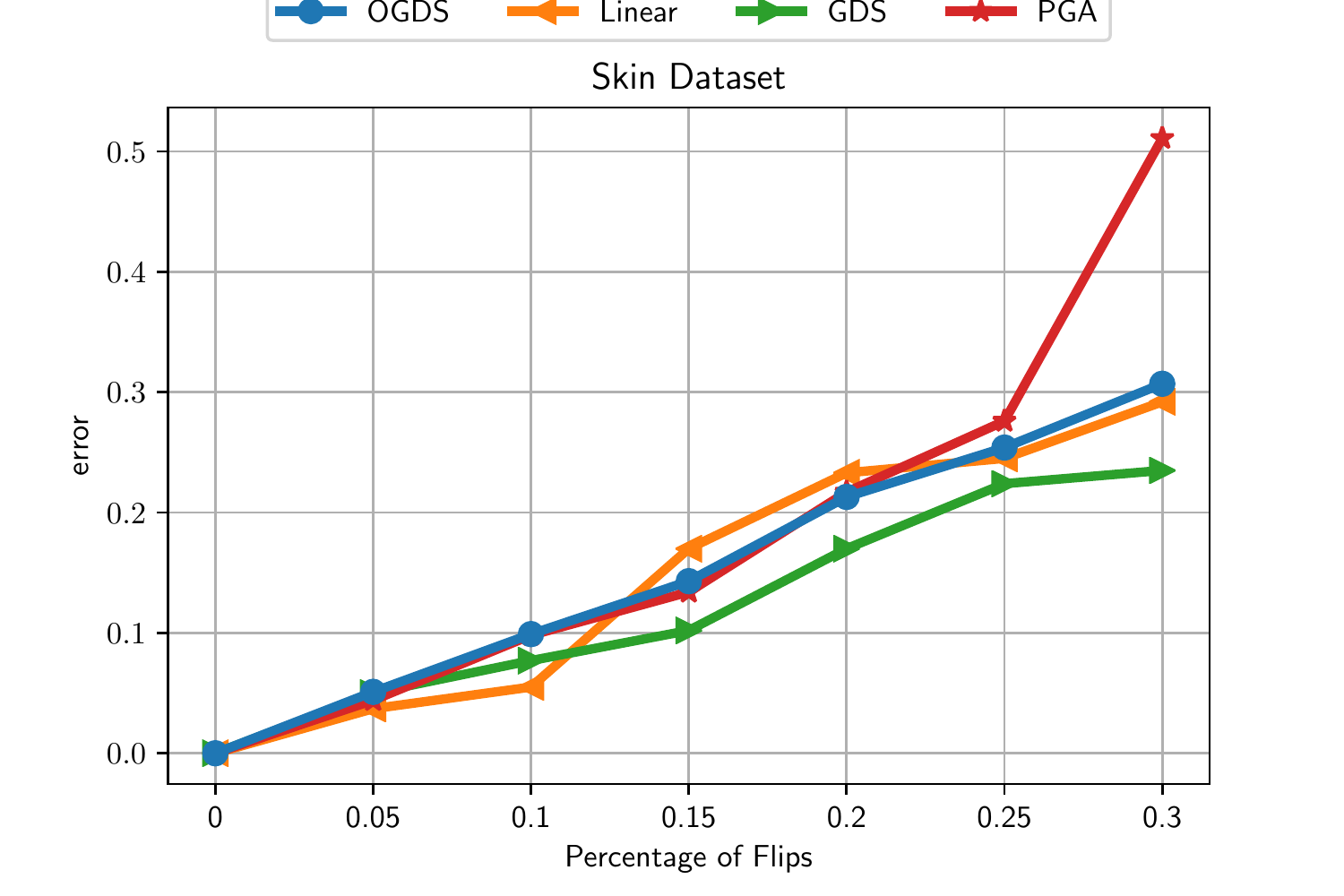} 
		\caption{LightGBM}  
		\label{fig:Lgbmskin}
	\end{subfigure}
	\begin{subfigure}[b]{0.24\textwidth}
		\centering
		\includegraphics[width=1\linewidth]{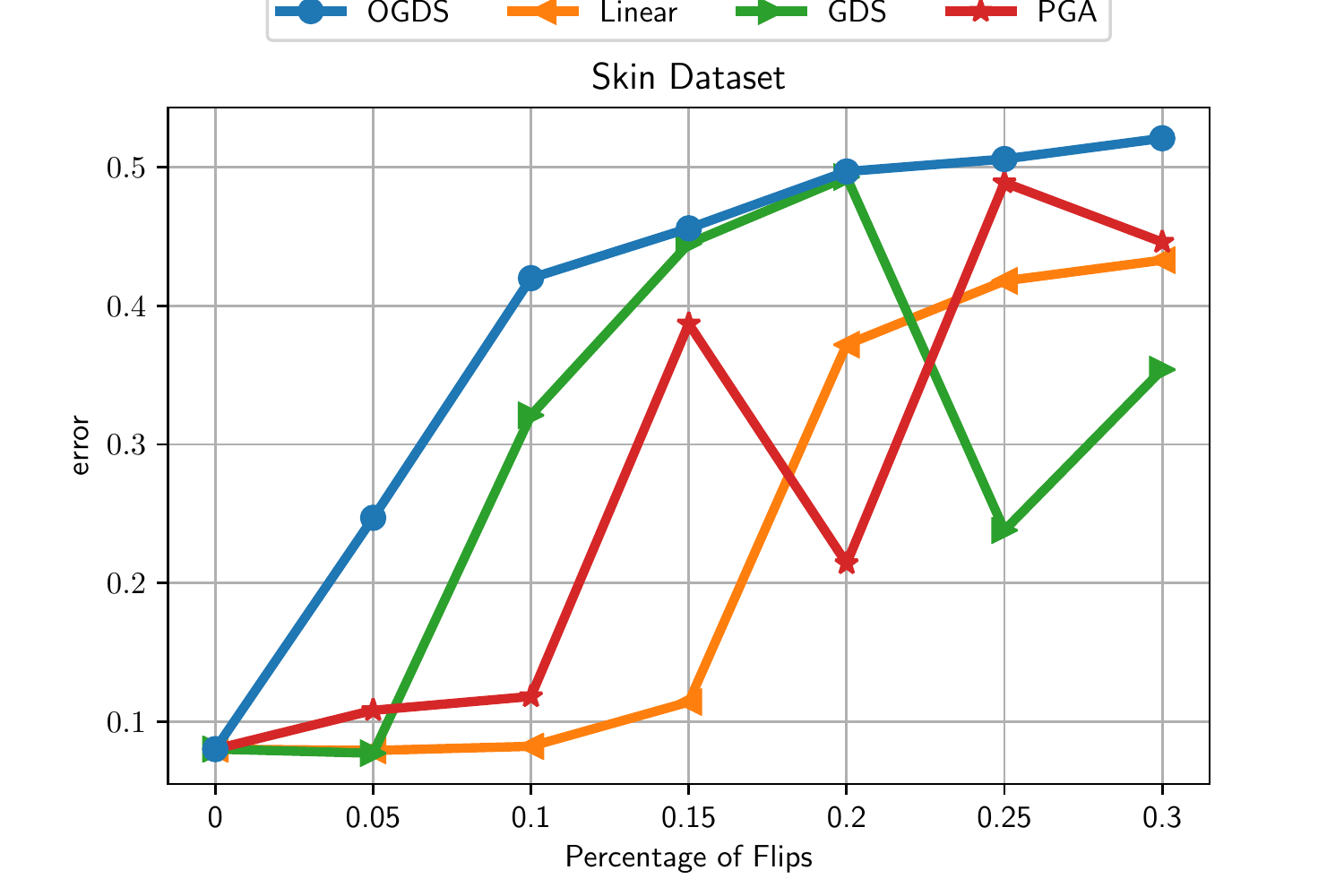} 
		\caption{Naive Bayes}  
		\label{fig:NBskin}
	\end{subfigure}
	\begin{subfigure}[b]{0.24\textwidth}
		\centering
		\includegraphics[width=1\linewidth]{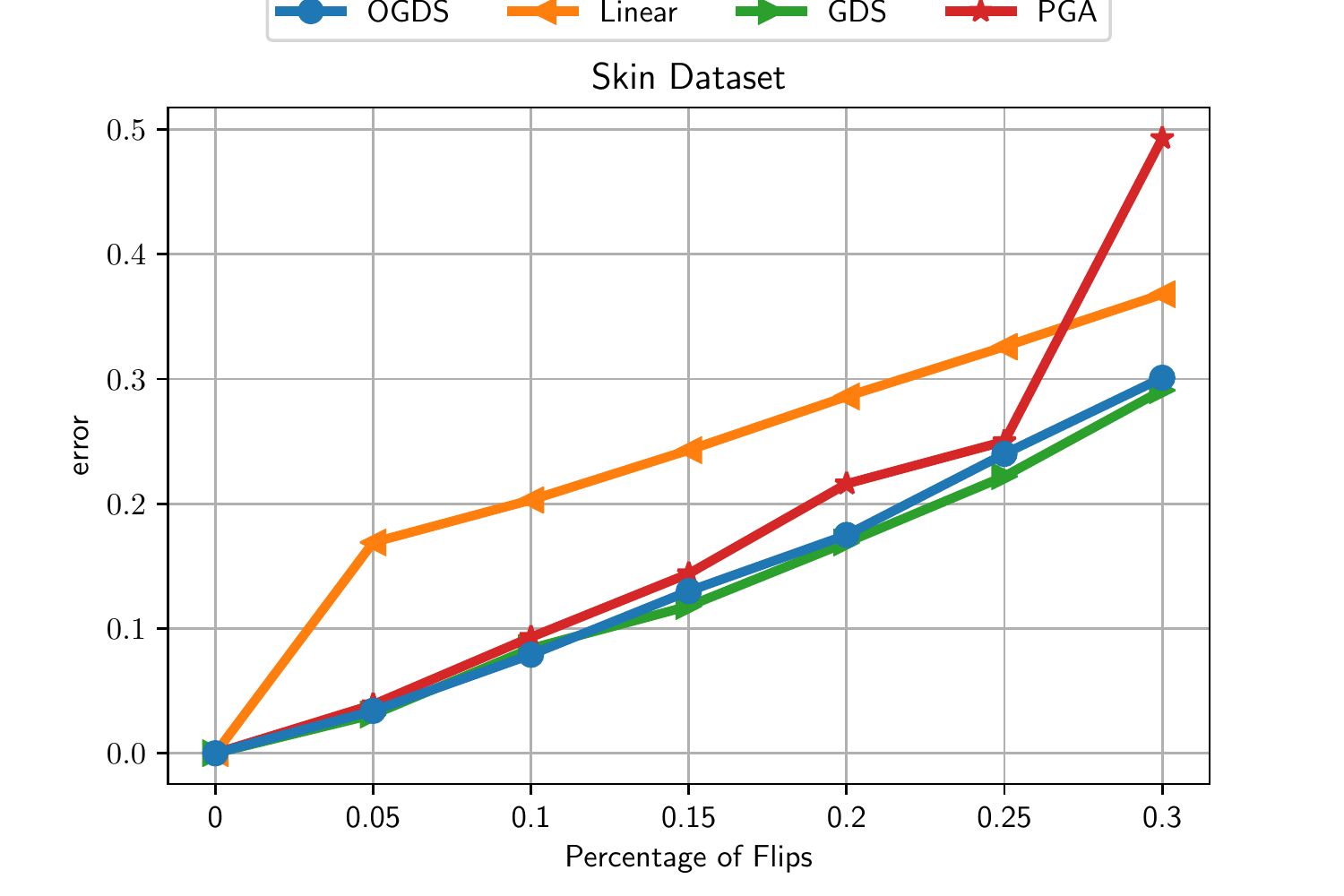} 
		\caption{KNN}  
		\label{fig:knnskin}
	\end{subfigure}
	\caption{Error rate of classifiers under various adversarial budgets $\mathcal{B}$ (0\% (Original Accuracy), 5\%, 10\%, 15\%, 20\%, 25\%, 30\%). Each row represents the result with Wine, Banknote, EMBER, Australian, Spambase and Skin datasets respectively. x-axis is the percentage of contaminated labels and y-axis is the error rate of the classifiers.} 
	\label{fig:B vs Error} 
\end{figure}

We can see that an attack can significantly degrade the accuracy of a victim model, even if it was not designed for the same model. We attribute this to the fact that the filtering of dataset is done based on the influence of the data instances on the training process. The variance of the attacks against various surrogate models on a victim model is much lesser than the variance of the attacks against a surrogate model on various victim models. This shows that the OGDS attack has a good transferability.

We illustrate the performance of each strategy under different budgets in Figure \ref{fig:B vs Error}. Each of the sub-figures shows the result using different learning algorithms in multiple datasets. For comparing PGA with classifiers other than Logistic Regression, we consider the transferability of attack. For instance, we cannot have Naive Bayes as a surrogate classifier in PGA. Instead, we can analyze the performance of Naive Bayes on the poisoned data generated by PGA using a different surrogate model. 

As we can observe from Figure \ref{fig:B vs Error}, OGDS performs better than the baselines for Logistic Regression across datasets. For LightGBM classifier, all attack strategies have similar performances, except PGA, that clearly performs better on Skin dataset. For Naive Bayes, the gradient-based approaches OGDS/GDS outperform the baselines. For KNN, PGA performs better on all datasets except Wine, Ember and Skin. Thus, it is evident from the results that the gradient-based attack strategies outperform the baselines in most cases.

\subsubsection{\bf Analysis of Cost Function}\label{sec:cost}
We have an adversarial cost $c_{i}$ associated with each instance $\mathbf{x}_{i}$. The cost value can be any non-negative real number. One can model a cost function which assigns high cost values to influential data instances. Our intuition is that such a function would constrain the well trained instances to be in the set of data instances to be flipped. To the best of our knowledge, prior works do not have enough discussion on varying the adversarial costs for label contamination \cite{xiao2012adversarial}.

\begin{table}[t]
    \centering
    \caption{Cost function analysis on EMBER dataset. Cost can be $uniform$ (set to \emph{1}) or $varied$ (set to \emph{1} for large gradient instances and set to \emph{2} for small gradient instances). $count_{a}$ and $count_{b}$ are the number of large and small gradient instances respectively.}
    \label{tab:cost function}
    \begin{tabular}{lc@{\hspace{0.5cm}} c@{\hspace{0.5cm}} c@{\hspace{0.5cm}} c@{\hspace{0.5cm}} c}
    \hline
    \textbf{\textbf{Cost}} & \textbf{\textbf{Budget ($\mathcal{B}$)}} & \textbf{\textbf{Error}} & \textbf{\textbf{$\#$ Flips}} & \textbf{\textbf{$count_{a}$}} & \textbf{\textbf{$count_{b}$}} \\ \hline
    \emph{uniform} & 600 & 0.4805 & 600 & 14 & 586 \\
    \emph{varied} & 1200 & 0.4885 & 616 & 32 & 584 \\ \hline
    \emph{uniform} & 1200 & 0.5211 & 1200 & 39 & 1161 \\
    \emph{varied} & 2400 & 0.5165 & 1238 & 70 & 1165 \\ \hline
    \emph{uniform} & 1600 & 0.5080 & 1600 & 78 & 1522 \\
    \emph{varied} & 3200 & 0.5073 & 1657 & 112 & 1544 \\ \hline
    \end{tabular}
\end{table}

We consider a uniform cost function for our experiments. We conjectured that varying the costs has no effect on the quality of the solution (Refer Section \ref{sec:ogds}). Our analysis in Table \ref{tab:cost function} \& \ref{tab:cost function2} with cost functions for different budgets shows that the attacker can achieve similar error rates by scaling the budgets accordingly in a uniform cost setting.

\begin{table}[ht]
    \centering
    \caption{Cost function analysis on Australian dataset. Cost can be $uniform$ (set to \emph{1}) or $varied$ ({[}c$_{1}$, c$_{2}${]} set to large and small gradient instances respectively). $count_{a}$ and $count_{b}$ are the number of large and small gradient instances respectively.}
    \label{tab:cost function2}
    \begin{tabular}{@{}lc@{\hspace{0.5cm}} c@{\hspace{0.5cm}} c@{\hspace{0.5cm}} c@{\hspace{0.5cm}} c@{}}
    \toprule
    \textbf{Cost} & \textbf{Budget ($\mathcal{B}$)} & \textbf{Error} & \textbf{$\#$ Flips} & \textbf{$count_{a}$} & \textbf{$count_{b}$} \\ \midrule
    \emph{uniform} & 207 & 0.367 & 207 & 4 & 203 \\ \midrule
    \emph{varied {[}1, 2{]}} & 414 & 0.372 & 210 & 5 & 205 \\
    \emph{varied {[}1, 3{]}} & 624 & 0.372 & 211 & 5 & 206 \\
    \emph{varied {[}1, 4{]}} & 832 & 0.372 & 212 & 6 & 206 \\ \bottomrule
    \end{tabular}
\end{table}

We experimented on \emph{OGDS - LR} for different budgets keeping the cost \emph{uniform} and \emph{varied}. Uniform cost sets all $c_{i}$ to a constant whereas varied cost sets a smaller cost to the large gradient instances and a larger cost to the small gradient data instances. In Table \ref{tab:cost function} \& \ref{tab:cost function2} we can observe that both cost functions report similar error rates with a similar number of flips under different budgets. Also, for the same budget the choice of cost function affects the number of flips and the error rate ($\mathcal{B} = 1200$) in Table \ref{tab:cost function}. Moreover, one can observe that the number of small gradient data instances ($count_{b}$) in the chosen data for label contamination is higher than that of the large gradient instances ($count_{a}$). This aligns with our intuition of choosing a majority of instances with small gradients (refer sub section \ref{sec:prelim}).

\subsection{\bf Complexity Analysis}\label{sec:ComplexityAnalysis}
In this subsection, we are going to analyze the complexity of label contamination attacks. As the number of data instances for the training phase increases, the complexity of finding optimal instances to poison increases. Previous works on attacking support vector machines (ALFA) solved a linear programming problem on \emph{2n} variables and a quadratic programming problem in each iteration of the algorithm\cite{biggio2012poisoning}. Further, Zhao et al \cite{zhao2017efficient} compared their approach (PGA) with ALFA and pointed out that they only train a linear classifier (dual form) in each iteration. Our approach picks only \emph{k} = ((\emph{a+b})$\times$100\% $\times$ n) instances from \emph{n} training samples for further computation such that \emph{k} $\le$ \emph{n}. 

PGA-LR solves the dual form of the classifier. Logistic Regression can be achieved by solving either the primal or the dual optimization problem. The efficiency of dual solvers are better than primal except when the dataset has fewer features than the data points \cite{yu2011dual}. For such datasets, primal solvers have faster convergence and lower cost per iteration $\mathcal{O}(\#\text{data instances} \times \#\text{features})$ \cite{lin2007trust} than the dual counterparts $\mathcal{O}(\#\text{data instances} \times \#\text{Newton steps})$ \cite{yu2011dual}. PGA approach involves matrix multiplication ($\mathcal{O}(n^{2.373})$ \cite{le2014powers}) for the gradient computation which further increases their complexity. 

We consider a large training dataset with $\#\text{data instances} > \#\text{features}$. Our attack strategy involves solving a linear programming problem of \emph{2k} variables ($\mathcal{O}((2k)^{3.5}\times(k)^{2})$) \cite{megiddo1986complexity} and the primal problem (Logistic Regression) in each iteration. Hence, OGDS is effective for large training data compared to the baseline approaches. sGDS (Algorithm \ref{algo:sgds}) involves sorting a coefficient list ($\mathcal{O}(k log k$) and solving a primal problem in each iteration. Therefore, sGDS is more efficient than OGDS asymptotically. The running time analysis of PGA, OGDS and sGDS is tabulated in Table \ref{tab:algo analysis}.

\begin{table}[t]
    \centering
    \caption{Running time analysis of algorithms. \emph{I} is the number of iterations, $\checkmark$ in \emph{LP} denotes the algorithm involves solving a linear programming problem and problem denotes the type of problem solved by the classifier.}
    \label{tab:algo analysis}
    \begin{tabular}{@{}lc@{\hspace{0.5cm}} c@{\hspace{0.5cm}} c@{}}
    \toprule
    \textbf{Method} & \textbf{LP} & \textbf{Problem} & \textbf{Running Time (seconds (s))} \\ \midrule
    PGA & $\times$ & Dual & 192.97 \emph{s} $\times$ \emph{I} \\
    OGDS & $\checkmark$ & Primal$\diagup$Dual & 24.91 \emph{s} $\times$ \emph{I} + 14.05 \emph{s} \\
    \textbf{sGDS} & $\times$ & Primal$\diagup$Dual & \textbf{20.97} \emph{s} $\times$ \emph{I} + \textbf{14.05} \emph{s} \\ \bottomrule
    \end{tabular}
\end{table}

\section{Susceptibility Analysis} \label{sec:susceptibility}
The \emph{susceptibility} of a classifier can be assessed by analyzing the error rate on a validation set when it is under an attack. We analyze it based on the transferability of attack on the EMBER dataset. Demontis et al \cite{demontis2018securing} in their recent work provided insights on factors affecting the transferability of attack such as the complexity of the surrogate model and its alignment with the victim model. Moreover, the authors recommend choosing a surrogate model with the regularization level similar to that of the victim model. However, their work focuses on perturbing the input samples using a gradient-based optimization framework rather than contaminating the input labels. In some application domains such as malware detection, perturbing the data instance is difficult while preserving their functionality. Hence, the relatively inexpensive nature of the label contamination may encourage the attackers to carry out the data poisoning attack. 

Table \ref{tab:SM VM} shows the error rate of victim models (LR, K-NN, NB and LightGBM) against corresponding surrogate models (PGA - LR, OGDS - [LR, KNN, LGBM, NB]). The first row shows actual error rate of the victim models on the validation set that is when the models are trained with reliable datasets. PGA - LR is the PGA attack strategy using the LR model which is then transferred to the aforementioned victim models. OGDS - [LR, KNN, LGBM, NB] denotes the proposed attack strategy utilizing classifiers (LR, LightGBM, K-NN, NB) for training in step 5 and 11 of Algorithm \ref{algo:OGDS}. 

\begin{table}[ht]
    \centering
    \caption{Susceptibility of victim models for malware classifiers (EMBER dataset). The poisoned labels learned from the surrogate models (\textit{left most column}) are used to train the victim models (\textit{top row of the table}) and the error rates are tabulated. Budget $\mathcal{B}$ = 1200 (attacker can only flip 10\% of data).}
    \label{tab:SM VM}
    \begin{tabular}{@{}l@{\hspace{0.5cm}} c@{\hspace{0.5cm}} c@{\hspace{0.5cm}} c@{\hspace{0.5cm}} c@{}}
    \toprule
    \textbf{} & \textbf{LR} & \textbf{KNN} & \textbf{LGBM} & \textbf{NB} \\ \midrule
    Actual Error Rate & 0.411 & 0.185 & 0.022 & 0.472 \\ \midrule
    PGA - LR & 0.477 & 0.19 & 0.078 & 0.488 \\ \midrule
    \textbf{OGDS - LR} & \textbf{0.499} & \textbf{0.291} & \textbf{0.29} & 0.479 \\
    OGDS - KNN & 0.418 & 0.245 & 0.046 & 0.477 \\
    OGDS - LGBM & 0.482 & 0.187 & 0.053 & 0.461 \\
    OGDS - NB & 0.454 & 0.198 & 0.044 & \textbf{0.488} \\ \bottomrule
    \end{tabular}
 \end{table}

The poisoned dataset ($\mathcal{D}_{p}$) obtained from the attack is used to train the victim models and their error rate on the validation data is reported. We can observe that OGDS-LR has a higher error rate on the victim LR model than other attack models. Moreover, its transferability of attack to the LightGBM model is significantly higher than that of OGDS - LGBM attack on LGBM (identical surrogate and victim models). Transferability of attack using OGDS - LR (linear classifier) is noteworthy as it has a significant impact on the performance degradation of victim models. 

Due to the poor performance of LR and NB (almost 60\% accuracy), this attack strategy could not degrade the model performance significantly (less than 10\%). LightGBM shows impressive performance as a malware classifier in \cite{anderson2018ember}. The original classifier's performance and the increase in error rate (almost 30\%) using OGDS-LR concludes that LightGBM is highly susceptible to the proposed attack. The error rate of K-NN which has a good performance on the untainted dataset, increased from 18\% to 29\% using OGDS. 

We can also infer from the transferability of our attack that the majority of the state-of-art machine learning algorithms are more susceptible under data poisoning attacks. The attacks computed against the surrogate models have performed well on a majority of the victim models (Table \ref{tab:transferUCI}). We conjecture that this study helps the machine learning community to build learning algorithms which are less susceptible for the attacks considering the criticality of the applications.

\section{Summary and Remarks on Defenses} \label{sec:defense}
\textbf{Related works on defenses.} In the early days, whenever a classifier was defeated, the only solution to the problem was to retrain the classifier. Dalvi et al \cite{dalvi2004adversarial} viewed classification as a game between the classifier and the adversary, thereby created a heuristic based algorithm to produce a classifier that defended against the attacks made by the adversary. Biggio et al \cite{biggio2018wild} have a good survey of defensive mechanisms developed over years. Many encouraging mechanisms have been developed by employing concepts in game theory and robust statistics \cite{alfeld2017explicit,an2016stackelberg,zhu2019} that are stable theoretically. Defensive mechanisms can be categorized into reactive and proactive defenses where reactive defenses aim to counter past attacks while proactive defenses aim to prevent future attacks. 

Adversarial attacks do not have a clear-cut set of rules. An attacker can formulate an attack from scratch and execute it. The two main difficulties while developing defense strategies are that 1. the attacker's goals are hard to estimate 2. most of the defense strategies assume that the data they handle is clean whereas in reality gathering reliable labeled information is expensive considering the human intervention involved.

Previous studies have shown that data sanitization can be used for mitigating the data poisoning attack \cite{steinhardt2017certified,cretu2008casting}. For example, if the given data point is far away from the distribution of the training data, then it is likely that the data point is poisoned. Besides, we can make the underlying model robust enough to prevent the attack. Training multiple classifiers and creating robust examples for the training phase are some of the strategies used to improve the robustness of the classifier \cite{biggio2011bagging,papernot2016distillation}. Apart from that, Zhao et al \cite{zhao2017efficient} discuss identification of the attack point and its adjacent points as one defensive mechanism. They also study game models for developing more secure learning algorithms.
Moreover, transfer attacks gained growing interest in the recent past. In a recent study, Shumailov et al. \cite{shumailov2019sitatapatra} discuss blocking of transferability of attacks against neural networks in computer vision applications.

\textbf{Possible future direction on defenses.} The learning capability of a model depends on the method used to convert raw features to model features. For example, Anderson et al \cite{anderson2018ember} used a combination of techniques such as hashing trick and histogram representation for creating suitable malware classifier features. Given that the victim model is a black-box to the attacker, he would have to choose a suitable feature conversion procedure for being able achieve his goal. Hence, we speculate that the choice of methods for raw feature conversion may affect the defensive capability of a classifier.

When the dataset is huge, defensive mechanisms such as training multiple classifiers or generating adversarial training points to make robust classifiers are expensive procedures in terms of cost and complexity. Moreover, centralized systems used to store data results in increased exposure of sensitive data and can lead to many attacks that cause extensive damages. Blockchains could be a potential remedy for this problem. A blockchain is a system in which a record of transactions is  maintained across several computers that are linked in a peer-to-peer network. The data is stored across a network of nodes, instead of a single central node since a blockchain is decentralized \cite{zyskind2015decentralizing}. If the data is maintained in a peer-to-peer network and each node runs its own different `sanity check' algorithm, then the attacker has a real tough job at hand. Although blockchain is not immune to problems, it would be interesting to see if we are able to use it to maximum advantage and create a defensive mechanism based on it.

\section{Discussion} \label{sec:discussion}
The extended use of data from external sources for training the machine learning models increases the risk of working with data corrupted by a malicious actor. We introduced a new label contamination strategy: \emph{gradient-based data subversion attack} for the data poisoning. This study complements the existing landscape of data poisoning attacks and helps us to proactively identify the threats of the machine learning system under design and devise countermeasures against them.

As discussed in the recent study of stronger data poisoning attack \cite{koh2018stronger}, the power of an attacker is restricted when she can only flip the labels rather than modifying the input. Most of the data poisoning attacks are focused on input perturbation (generating the adversarial examples) on the other hand, label contamination has not gathered similar interest. Our work is focused on the effective flipping of data points which will be feasible in the cases of applications such as malware detection, object recognition, intrusion detection where the functionality of the data samples has to be preserved.

Our approach produces the poisoned version of the dataset that maximally degrades the victim model. Here, the adversary can access training data and more specifically in our threat model the adversary can contaminate only the labels in training data. Further, the process of identifying the data points to flip is independent of the knowledge of the victim model. We speed up the process by reducing the search space of data instances, based on sampling technique proposed in \cite{ke2017lightgbm}. Data instances with different gradients play different roles in the computation of information gain.

Although our proposed attack strategies shows improvement over existing attacks in majority of the datasets, a more rigorous study on \emph{non-linear learning algorithms} is required. Hence, the main limitation to this work is that we have not experimented our approach across deep learning algorithms, to understand the model agnostic nature of our attack.  Therefore, we would like to conduct a systematic investigation of label contamination strategies against deep networks in the near future. Further, recent study shows the robustness of the statistical inferences from corrupted data \cite{zhu2019}. Motivated by this, we consider exploring the variation in robustness of the learning algorithm under our attack as a potential next step.

Previous work exploited the data poisoning framework to analyze the vulnerability of feature selection algorithms \cite{xiao2015feature}. Apart from the discussed feature selection algorithms such as LASSO, ridge regression and the elastic net. It would be interesting to analyze the impact of label contamination on feature selection approaches such as tree-derived feature importance. The explanation of the attack on non-differentiable algorithms would be another interesting direction. Additionally, one could study the effect of label noise added based on the interpretability of non-differentiable algorithms. The next natural step is to extend the current attack strategy for the multi-class classification problem and also for an online learning setting.

\section{Conclusion} \label{sec:conclusion}
The security of machine learning algorithms poses a serious threat despite their success in multiple domains. In this work, we studied the data poisoning attack on training data for binary classifiers. In particular, we focused on the label contamination attack. We demonstrated that our attack algorithms (GDS, OGDS, sGDS) based on the gradients of a differentiable convex loss function (residual errors) outperforms the baselines. We explored how the choice of cost function affect the quality of the solution and discussed the transferability property of our attack on multiple datasets. Further, we analyzed the susceptibility of the state-of-art machine learning algorithms under our proposed attack. We also discussed the possible defense strategies to prevent the poisoning attacks.

\bibliographystyle{splncs04}
\bibliography{data_input_poison_bib}

\end{document}